\documentclass[preprint,12pt]{elsarticle}

\usepackage{etoolbox}
\patchcmd{\subsubsection}{\itshape}{\bfseries}{}{}
\patchcmd{\subsection}{\itshape}{\bfseries}{}{}

\usepackage{amssymb}
\usepackage{amsmath}
\usepackage{booktabs}
\usepackage{bm}
\usepackage{graphicx}
\usepackage{tabularx}
\usepackage{makecell}
\usepackage{multirow}
\usepackage{lineno}

\journal{Advanced Engineering Informatics}
\begin{document}

\begin{frontmatter}

\title{A physics-coupled graph neural operator for mid-plane six-component stress tensor field prediction in concrete penetration}

\author[1]{Wenpu Du}
\author[1]{Peng Zhou}
\author[1]{Yunlong Xia}
\author[1]{Sinuo Xin}
\author[1]{Congcong Zhang}
\author[1]{Boyang Zhang}
\author[1]{Yi Zhang}
\author[1]{Wenzheng Xu\corref{cor1}}

\cortext[cor1]{Corresponding author. E-mail address: xuwznuc@163.com}

\affiliation[1]{organization={School of Environment and Safety Engineering, North University of China},
	city={Taiyuan},
	postcode={030051},
	country={China}}

\begin{abstract}

The penetration-resistant design of concrete structures requires knowledge of the spatiotemporal evolution of the stress field within the target under projectile impact, whereas in engineering practice only the strength grade, the structural geometry and the projectile parameters can be determined, and the spatial arrangement of the aggregates is not accessible. Empirical formulae that take macroscopic parameters as input and data-driven surrogates that output scalars serve design screening, and mesoscale finite element models can resolve the response at the aggregate scale, but the cost of a single solution cannot support coverage of random aggregate samples at scale, and a unified framework that connects mesoscale heterogeneity with field-level prediction has not been established. To address this gap, a physics-coupled graph neural operator (PC-GNO) is proposed, which learns in function space the evolution of the mid-plane six-component stress tensor field of the target under a given impact velocity. The training data are generated by full-scale, aggregate-resolved explicit finite element models and comprise 359 cases, divided by impact velocity into the three levels of 100, 135 and 165 m/s, which correspond respectively to the three concrete compressive strengths of 14, 23 and 48 MPa; the credibility of the data source is verified by laboratory measurement data. The operator constructs the coupling strength of message passing on the graph from the impedance contrast on the two sides of the material interface, so that the physical prior enters the edge weights, and the propagation operator constrains the per-round update by spectral normalization. Under the single-step primary setting, the average accuracy of the three levels is 80.16\%, the 100, 135 and 165 m/s levels giving 91.45\%, 77.80\% and 71.22\% respectively, which are 27.81, 42.46 and 34.16 percentage points above the preceding-frame field of the same window; the number of learnable parameters of the operator is 20\,664, and under the same configuration of the 100 m/s level its accuracy is higher than the 87.9\% of the best configuration of the Fourier neural operator, whose number of learnable parameters is about 812 times that of the operator, and the 71.0\% of the closed-form linear regression. The operator outputs the complete six-component tensor field rather than a fixed engineering quantity, and strength criteria, damage criteria and energy indicators can be derived from one and the same set of weights. Its contribution lies in integrating the mesoscale simulation data resource, the field-level operator and the two-step verification chain into one whole, and in giving accuracy results through a cross-case evaluation. For the field-level response prediction of concrete penetration, this result can provide a data-driven route that is connected after the computation, and its conditions of applicability are defined by the working conditions covered by the training and the evaluation.

\end{abstract}

\nolinenumbers
\begin{highlights}
\item PC-GNO predicts the evolving mid-plane six-component stress tensor field.
\item Single-step primary setting: 80.16\% mean accuracy, above the preceding frame.
\item 20\,664 parameters reach it; the Fourier operator needs 812 times more.
\item One set of weights yields strength, damage and energy criteria alike.
\item Verified against 134 m/s tests: 2.51 ms deceleration, 133.0 $\pm$ 1.2 mm crater.
\end{highlights}
\linenumbers


\end{frontmatter}

\section{Introduction}
\label{sec:1}

The penetration response of concrete is a central concern of protective engineering and impact-resistant design; from protective fortifications and nuclear containment shells to civil crash-resistant structures, the penetration resistance directly determines the protective capacity and the safety margin of a structure. Such a design faces a fundamental input difficulty: the only quantities that can be determined in engineering practice are the concrete strength grade, the structural geometry and the parameters of the incoming projectile, whereas the spatial arrangement of aggregates within the actual structure cannot be obtained and is difficult to determine by conventional inspection. Aggregates are the carrier of the heterogeneity of concrete: under the impact of a high-velocity projectile, the failure process proceeds in sequence through the mutually coupled stages of nose erosion, cratering at the target surface, tunnel-region extension and back-face scabbing \cite{ref1,ref2,ref3}, and the failure mode at each stage is influenced by the mesoscale structure. The role of the mesoscale structure exhibits a clear hierarchy: at the same strength grade and the same impact velocity, the variation of the aggregate spatial arrangement exerts a limited influence on macroscopic penetration indices, its effect being mainly manifested in the local stress concentrations around the aggregates, in the shape and attenuation of the stress wave front and in the field-level spatial distribution of damage. What the design therefore needs to answer is the spatiotemporal evolution of the mid-plane stress field at a given impact velocity, rather than the computed result of one particular realization of the aggregate arrangement.

Empirical formulas and the cavity expansion approximation, whose output is the penetration depth, give explicit estimates in terms of macroscopic parameters such as the target strength, the projectile mass, the nose shape and the impact velocity, and the computational cost is low, which facilitates rapid engineering estimation \cite{ref4,ref5,ref6,ref7,ref8}; however, the material is represented by averaged quantities such as the volume fraction and the mean strength, so that information on the aggregate spatial arrangement is not retained and the field-level stress response cannot be given. Data-driven surrogates replace high-cost solution procedures at a lower cost: artificial neural networks, ensemble learning and interpretable machine learning have been widely used for the regression prediction of scalar penetration indices \cite{ref9,ref10,ref11,ref12,ref13,ref14}, and the generalization capability has been examined with respect to parameters such as the projectile configuration, the target strength and the randomness of protective structures. In the more than twenty representative studies examined in the present work, such methods take a uniform form: the input consists of four to eight design parameters that are determined before the simulation begins, the output consists of one to three scalars including the penetration depth, the crater depth and diameter, the residual velocity and the ballistic limit; the data volume ranges from tens to hundreds of records, and the accuracy index is the mean absolute percentage error or the coefficient of determination obtained on a random split of the same dataset, in most cases not computed on independent real tests. Such work undertakes a design-screening function, and its input is determined before the computation.

Mesoscale finite element simulation models the aggregates and the mortar matrix as separate phases and can resolve the stress concentrations and the crack propagation around the aggregates \cite{ref15,ref16,ref17,ref18,ref19}; the modulation of the penetration response and damage by the aggregate volume fraction, the particle size and the spatial distribution has been revealed; and combinations with machine learning have also been developed for mesoscale impact response prediction and uncertainty quantification \cite{ref22,ref23,ref24}. Such methods can resolve the local field-level response, yet a single explicit solution is expensive under large deformation and element deletion \cite{ref20,ref21}, so that random aggregate samples cannot be covered at scale; the output is moreover dominated by scalar penetration indices, and the full-field stress tensor has rarely been systematically organized as a reusable dataset.

Work that takes a complete physical field rather than a scalar as the learning target remains uncommon. The PGN proposed by Li et al. uses a graph network to learn the spatial and temporal field quantities of the process in which a rigid projectile penetrates a target plate, and gives predictions of the displacement field, the stress field and the strain field \cite{ref25}. GAT-ImpactNet proposed by Gao et al. uses a graph attention network to predict the full-field equivalent plastic strain within an ultra-high performance concrete target, where the training, validation and test sets are taken from different time intervals of one and the same simulation; that study states that this setting cannot be generalized across physical configurations, and its accuracy claim is made on scalar engineering quantities \cite{ref26}. For the spatiotemporal prediction of structural impact responses, attempts have also been made to give the full-field response with a dynamic graph network \cite{ref27}. These two classes of work differ in task type from the design screeners described above: the input is the history field produced during the solution process, which becomes available only after the computation, and the function is to give the prediction of the subsequent field without a repeated solution, so that the accuracy figures of the two are not directly comparable. The differences of interest in the present work lie in two respects. The first is the manner of validation: the field-level work described above is evaluated within one and the same configuration, whereas usability in the engineering sense requires the operator to reproduce the field under a cross-case setting, that is, without dependence on a particular aggregate arrangement and with the same set of input interfaces giving predictions at different impact velocity levels. The second is the output dimensionality: derived quantities such as the equivalent plastic strain are single-component scalar fields, whereas the prediction of the mid-plane six-component stress tensor requires the simultaneous characterization of the coupling among the six components and of the strong discontinuities near the shock front, so that the dimensionality and stiffness problems are markedly more severe. Neural operator methods are characterized by learning operator mappings in a function space and have been widely applied to partial differential equation solving and parametric field prediction \cite{ref28,ref29,ref30}; within the scope examined in the present work, their use for the mid-plane six-component stress tensor field of concrete penetration has not been reported.

To address this gap, a physics-coupled graph neural operator (PC-GNO) is proposed, in which aggregate-resolved mesoscale simulation data serve as the training source and the time-varying evolution of the mid-plane six-component stress tensor field at a given impact velocity is learned in a function space. The training data were generated with a full-scale aggregate-resolved LS-DYNA model and comprise 359 cases, which are divided by impact velocity into the three levels of 100, 135 and 165 m/s, corresponding respectively to the concrete compressive strength levels of 14, 23 and 48 MPa; the three levels are modelled independently of each other, so that the generality of the operator is examined under multiple impact velocities and multiple concrete compressive strengths. The credibility of the simulation data platform is verified by laboratory ballistic penetration tests. Two groups of tests were completed on 23 MPa target plates with the same ballistic apparatus: in the first group, penetration was performed with a projectile of 340 mm total length at an initial velocity of 134 m/s, the measured deceleration duration being 2.51 ms, and under the same projectile total length, impact velocity and target strength as this group, the deceleration time history given by the simulation is consistent with the measured oscillatory curve in magnitude and duration; in the second group, penetration was performed with a projectile of 235 mm total length at an initial velocity of 135 m/s, and the crater depth measured by three-dimensional laser scanning is 133.0 $\pm$ 1.2 mm. A two-step chain is thereby formed: the laboratory measurement data are used to verify the explicit finite element simulation and the data source, the simulation data are used to train the field-level operator, and the operator is then subjected to cross-case validation.

The accuracy is defined by the relative $L_2$ error, and the accuracy percentage is defined as one minus the relative $L_2$ error, multiplied by one hundred; the relative $L_2$ error is computed frame by frame and then averaged over the window, with the norm of the solver ground-truth field over the same window as the denominator. Under the single-step primary setting, the average accuracy over the three levels is 80.16\%, the accuracies at 100, 135 and 165 m/s being 91.45\%, 77.80\% and 71.22\%, which exceed the preceding-frame field over the same window by 27.81, 42.46 and 34.16 percentage points respectively. The operator has 20\,664 learnable parameters; under the same configuration at the 100 m/s level, its accuracy exceeds 87.9\%, the value of the best configuration of the Fourier neural operator, whose learnable parameter count is about 812 times as large, and also exceeds 71.0\% of the closed-form linear regression.

The contributions are threefold. First, a physics-coupled graph neural operator is proposed, which learns in a function space the time-varying evolution of the aggregate-resolved mid-plane six-component stress tensor field, and accuracy results are given by cross-case evaluation. Second, a mid-plane six-component stress tensor dataset is constructed, which is full-scale and aggregate-resolved and spans three impact velocity levels and three levels of concrete compressive strength, as the data resource supporting the operator. Third, a two-step verification chain composed of laboratory measurement data and explicit finite element simulation is given, and the scope of application of each link in the chain is stated. Chapter~\ref{sec:2} presents the mesoscale numerical modelling, Chapter~\ref{sec:3} the verification of the laboratory measurement data and the simulation data source, Chapter~\ref{sec:4} the dataset, Chapter~\ref{sec:5} the field-level operator, Chapter~\ref{sec:6} the field-level results and comparisons together with the derivation and the scope of application of the field-level mechanism quantities, and Chapter~\ref{sec:7} the conclusions.

\section{Modelling}
\label{sec:2}

This chapter presents the construction of the aggregate-resolved mesoscale numerical model. The model serves as the data generation platform and provides training data for the dataset in Chapter~\ref{sec:4} and the field-level operator in Chapter~\ref{sec:5}. The projectile and target geometry and the material models, the random generation of aggregates, the explicit solution settings and the field extraction are presented in order.

\subsection{Projectile and target geometry and material models}
\label{sec:2_1}

The numerical model consists of a conical-nose projectile and a cylindrical concrete target plate, both modelled in the mm-g-ms-N-MPa unit system, in which 1 N = 1 g$\cdot$mm/ms$^2$, 1 MPa = 1 N/mm$^2$, and densities are expressed in g/mm$^3$.

The projectile is a hollow body with a conical nose, with a diameter of 45 mm, a total length of 235 mm, a nose full cone angle of 35.5$^\circ$ and a nose length of 71.21 mm. A coaxial cavity is provided inside the projectile, with a cavity diameter of 15 mm, a cavity length of 49.85 mm and a wall thickness of 15 mm; the projectile mass is 2.2414 kg. The projectile is described by the rigid-body constitutive model MAT\_RIGID, with a density of $7.85\times10^{-3}$ g/mm$^3$, an elastic modulus of 210000 MPa and a Poisson ratio of 0.30, and is discretized into 1840 elements. The kinematic quantities of the projectile are given by the rigid-body motion output of the solver. The projectile geometry of the dataset in the present work is modelled uniformly according to the values listed above, whereas the verification cases in Chapter~\ref{sec:3} are established separately according to the total length of the experimental projectile, and the difference between the two is given in Section~\ref{sec:3_2}.

\begin{figure}[!ht]
	\centering
	\includegraphics[width=\linewidth]{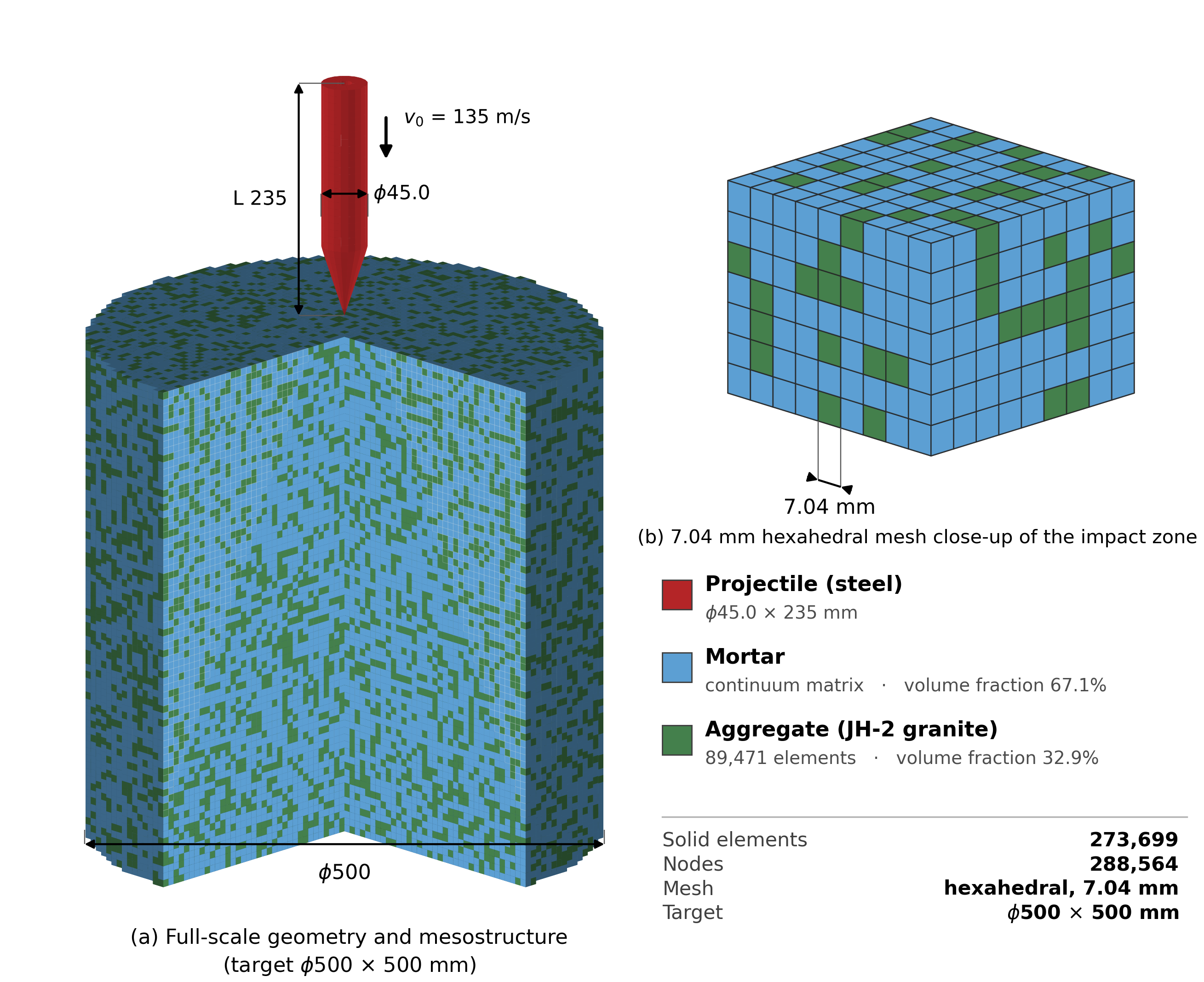}
	\caption{Geometry and mesh of the full-scale model. (a) The geometrical relation between the projectile and the cylindrical target plate and the coarse aggregates randomly distributed within the target; (b) the local mesh refinement in the impact region. The two panels are taken from one and the same rendering, the projectile, the mortar matrix and the coarse aggregates being distinguished by different colours, and the dimension lines give the projectile diameter, the projectile length and the initial projectile-target spacing.}
	\label{fig:1}
\end{figure}

The target plate is a cylinder of 500 mm diameter and 500 mm height, which is modelled with a two-phase mesoscale structure composed of a mortar matrix and coarse aggregates and is discretized into 271859 seven-millimetre hexahedral elements. For the numerical simulation of concrete target penetration, the selection and the evaluation of constitutive models have been studied systematically \cite{ref31,ref32,ref33}. The mortar matrix adopts the von Mises elastic-perfectly plastic constitutive model (MAT\_PLASTIC\_KINEMATIC, MAT\_003), with a density of $2.2\times10^{-3}$ g/mm$^3$, an elastic modulus of 10000 MPa and a Poisson ratio of 0.20, and the yield strength is assigned separately for each velocity level; the concrete compressive strengths of the three target plates are 14, 23 and 48 MPa respectively. The three levels vary simultaneously in the two dimensions of impact velocity and concrete compressive strength, so that the generality of the operator is examined under multiple impact velocities and multiple concrete compressive strengths. The coarse aggregates are described by the Johnson-Holmquist ceramic constitutive model (MAT\_JOHNSON\_HOLMQUIST\_CERAMICS, MAT\_110) \cite{ref34}, with a density of $2.7\times10^{-3}$ g/mm$^3$. The projectile and target geometry, the two-phase mesoscale structure and the mesh discretization are shown in Figure~\ref{fig:1}. The material and solution parameters are summarized in Table~\ref{tab:1}.

\subsection{Random generation of aggregates}
\label{sec:2_2}

\begin{figure}[!ht]
	\centering
	\includegraphics[width=\linewidth]{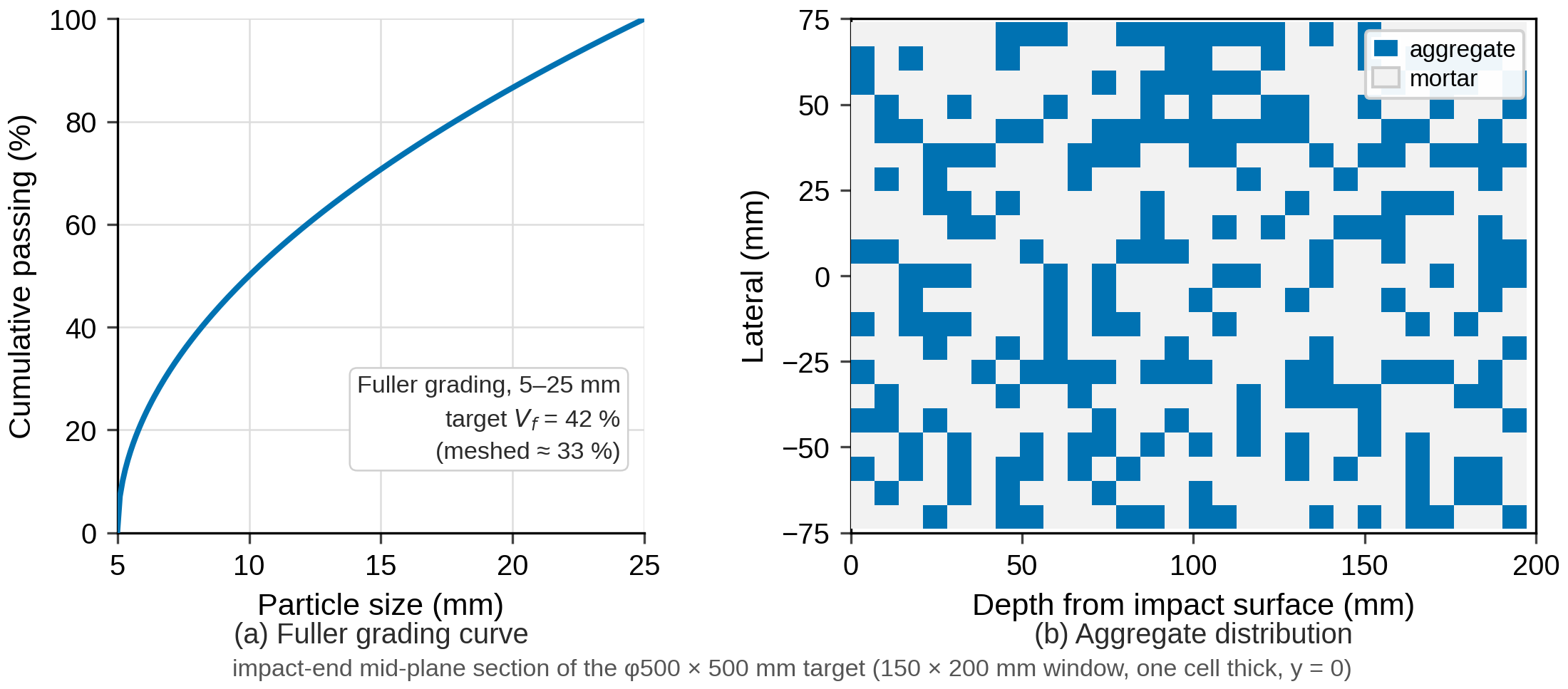}
	\caption{Aggregate gradation curve and geometric scatter. (a) The cumulative passing curve of the Fuller continuous gradation, with a particle size range of 5 to 25 mm and a target sphere volume fraction of 42\%; (b) the aggregate distribution generated by random placement on the mid-plane section at the impact face, the aggregate and mortar phases being coloured separately; the case is taken from the 100 m/s level with the number 1001, and the section is the layer of elements containing the target axis, with a lateral window of 150 mm and a depth of 200 mm; the aggregate proportion shown in the figure is an element-level value after meshing and is lower than the target sphere volume fraction.}
	\label{fig:2}
\end{figure}

The aggregate arrangement is generated by a random placement algorithm. The gradation follows the Fuller continuous gradation, the particle size range is taken as 5 to 25 mm, the target sphere volume fraction is 42\%, and the placement domain is the cylindrical region occupied by the target body, with a diameter and a height of 500 mm. The placement proceeds from large to small particle sizes, with the overlap between spheres checked particle by particle until the target volume fraction is reached. The aggregate positions are controlled by random seeds; within one velocity level the cases differ only in the random seed while the material and solution settings are kept identical, so that aggregate arrangements that are geometrically independent of each other and physically identically distributed are produced. The gradation curve and the aggregate distribution on the mid-plane section at the impact face are shown in Figure~\ref{fig:2}.

\subsection{Explicit solution settings}
\label{sec:2_3}

The penetration process is solved with the explicit dynamics program LS-DYNA \cite{ref35}. Both the target plate and the projectile use single-point integration hexahedral elements (ELFORM=1) with an element size of 7.0 mm. An eroding surface-to-surface contact (CONTACT\_ERODING\_SURFACE\_TO\_SURFACE) is used between the projectile and the target plate, and all faces of the target plate are free boundaries. The initial velocity is applied to the projectile with INITIAL\_VELOCITY\_GENERATION, and is set to 100, 135 and 165 m/s for the three levels respectively. The time integration adopts the explicit central difference scheme, with a time step scale factor TSSFAC of 0.9 and a mass scaling parameter DT2MS of $-1\times10^{-4}$ ms; the hourglass control is of type 6 with a coefficient of 0.10; the bulk viscosity coefficients are taken as 1.5 and 0.06. The termination time is set to 2.000 ms, the field snapshot interval is 0.025 ms, and 82 frames are output for each case.

\subsection{Field extraction and normalization}
\label{sec:2_4}

\begin{figure}[!ht]
	\centering
	\includegraphics[width=\linewidth]{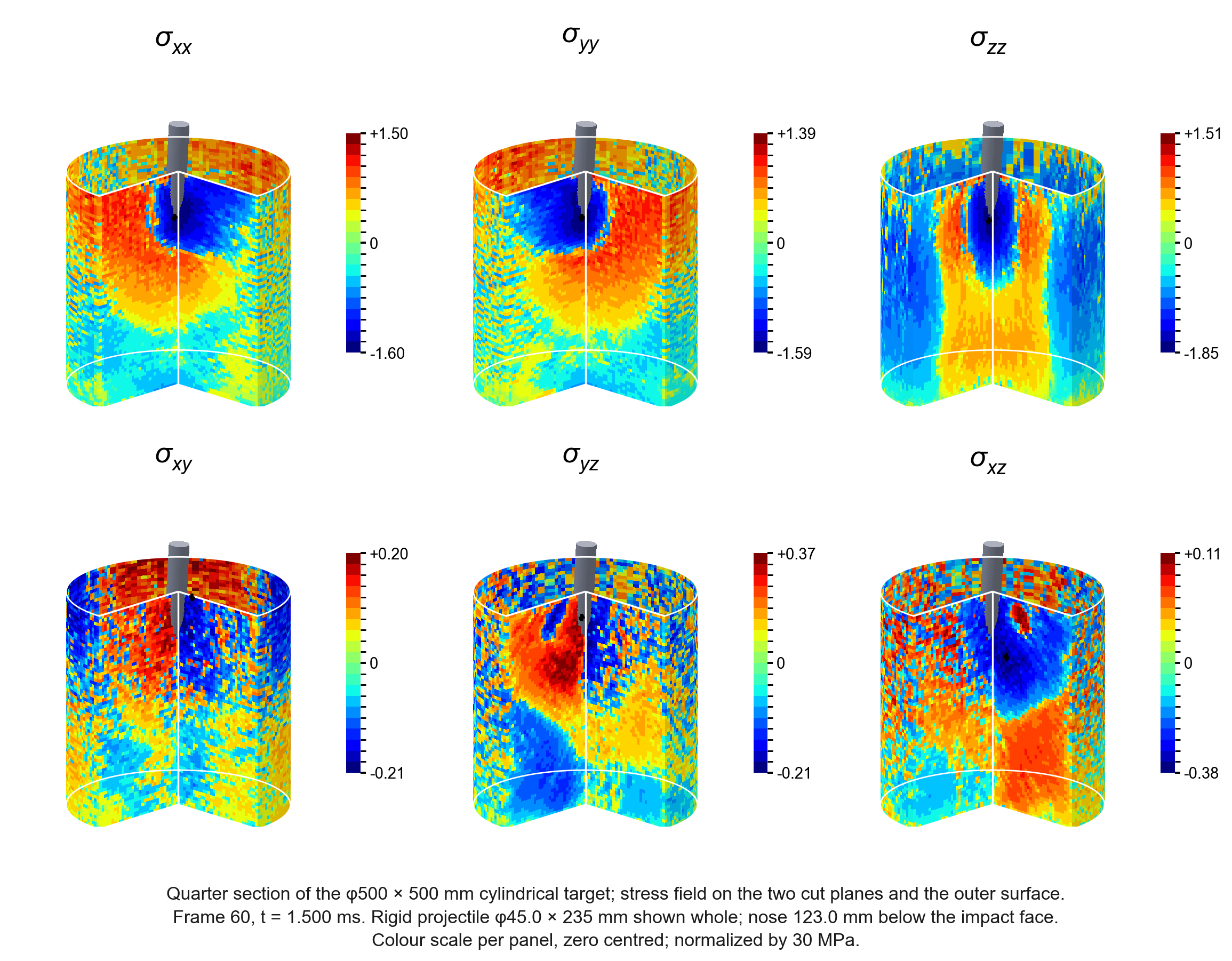}
	\caption{Six-component stress snapshots of a representative case. The six panels are, in order, $\sigma_{xx}$, $\sigma_{yy}$, $\sigma_{zz}$, $\sigma_{xy}$, $\sigma_{yz}$ and $\sigma_{xz}$, and are taken from the mid-plane field of one and the same case and frame. Each component is presented as a rectangular block, the range of each panel being given separately with an independent colour bar, and the colour bars are displayed with a symmetric power-law spacing; the stress values are the normalized values.}
	\label{fig:3}
\end{figure}

The field quantities are derived from the element stresses output by LS-DYNA and are resampled to a regular $64\times64$ cross-section grid by linear assignment with bilinear weight normalization: in the depth direction Z the coverage starts from the impact face and spans 0 to 500 mm, in the lateral direction X the coverage spans $-250$ to 250 mm, the grid spacing in both directions is 7.8125 mm, and the grid convention is the element centre. The stress field is stored as a four-dimensional array of $[6,64,64,82]$, in which the first dimension is the stress component, ordered as $\sigma_{xx}$, $\sigma_{yy}$, $\sigma_{zz}$, $\sigma_{xy}$, $\sigma_{yz}$ and $\sigma_{xz}$, the two middle dimensions are the spatial discretization of the cross-section, and the last dimension is the time frame. All stress values are normalized in the export layer by division by 30 MPa, and the physical values are recovered by multiplication by this factor. The volume fraction field is stored separately as a three-dimensional array of $[64,64,64]$, whose three dimensions correspond to the Z, X and Y directions respectively. The six-component stress snapshots of a representative case are shown in Figure~\ref{fig:3}.

\begin{table}[!ht]
\centering
\caption{Material and solution parameters}\label{tab:1}
\footnotesize
\renewcommand{\arraystretch}{0.95}
\setlength{\tabcolsep}{3pt}
\begin{tabularx}{\textwidth}{>{\raggedright\arraybackslash}p{1.4cm} >{\raggedright\arraybackslash}p{3.4cm} >{\raggedright\arraybackslash}p{3.0cm} >{\raggedright\arraybackslash}X}
\toprule
Phase & Constitutive model & Parameter & Value \\
\midrule
Projectile & MAT\_RIGID & Density / elastic modulus / Poisson ratio / mass & $7.85\times10^{-3}$ g/mm$^3$ / 210000 MPa / 0.30 / 2.2414 kg \\
\midrule
Mortar matrix & MAT\_\allowbreak PLASTIC\_\allowbreak KINEMATIC (MAT\_\allowbreak 003) & Density / elastic modulus / Poisson ratio & $2.2\times10^{-3}$ g/mm$^3$ / 10000 MPa / 0.20 \\
\midrule
Target & --- & Concrete compressive strength (three levels) & 14 / 23 / 48 MPa \\
\midrule
Coarse aggregate & MAT\_\allowbreak JOHNSON\_\allowbreak HOLMQUIST\_\allowbreak CERAMICS (MAT\_\allowbreak 110) & Density & $2.7\times10^{-3}$ g/mm$^3$ \\
\midrule
Aggregate geometry & Fuller gradation & Particle size range / target sphere volume fraction / placement domain & 5 to 25 mm / 42\% / $\phi$500$\times$500 mm cylinder \\
\midrule
Mesh & ELFORM=1 single-point integration hexahedron & Element size / target element count / projectile element count & 7.0 mm / 271859 / 1840 \\
\midrule
\multirow{2}{1.4cm}{Solution control} & \multirow{2}{3.4cm}{Explicit central difference} & TSSFAC / DT2MS / hourglass / bulk viscosity & 0.9 / $-1\times10^{-4}$ ms / type 6, 0.10 / 1.5 and 0.06 \\
& & Output & Termination time / snapshot interval / frame count: 2.000 ms / 0.025 ms / 82 \\
\midrule
\multirow{2}{1.4cm}{Field quantity} & \multirow{2}{3.4cm}{Resampling grid} & Grid spacing / coverage / direction & 7.8125 mm / Z 0 to 500 mm, X $-250$ to 250 mm / element centre \\
& & Storage & $[6,64,64,82]$ / $\sigma_{xx}$ $\sigma_{yy}$ $\sigma_{zz}$ $\sigma_{xy}$ $\sigma_{yz}$ $\sigma_{xz}$ / linear normalization (see Section~\ref{sec:4_2}) \\
\bottomrule
\end{tabularx}
\end{table}

\section{Real experiment}
\label{sec:3}

\subsection{Overview of the tests}
\label{sec:3_1}

The experimental data of this chapter are used to verify the mesoscale numerical model of Chapter~\ref{sec:2} and the data source generated by that model. The tests were completed on the same ballistic apparatus, and the target plates of the two groups of tests are both cylinders of 500 mm diameter and 500 mm height, and the compressive strength of both is 23 MPa, as determined by measurements on standard specimens prepared from the same batch as the target plates. The influence of the diameter and the boundary conditions of a concrete target on the projectile deceleration and the penetration depth has been studied experimentally \cite{ref36}, and the dimensions of the target plate are determined by the existing configuration of the test apparatus. Both groups of tests consist of normal penetration of the target plate by the projectile and are distinguished by the total projectile length: the projectile of the first group has a total length of 235 mm and an impact velocity of 135 m/s, and the target plate was subjected to three-dimensional laser scanning after penetration in order to reconstruct the crater; the projectile of the second group has a total length of 340 mm and an impact velocity of 134 m/s, and an acceleration sensor was installed inside the projectile to record the deceleration during penetration.

\begin{figure}[!ht]
	\centering
	\includegraphics[width=\linewidth]{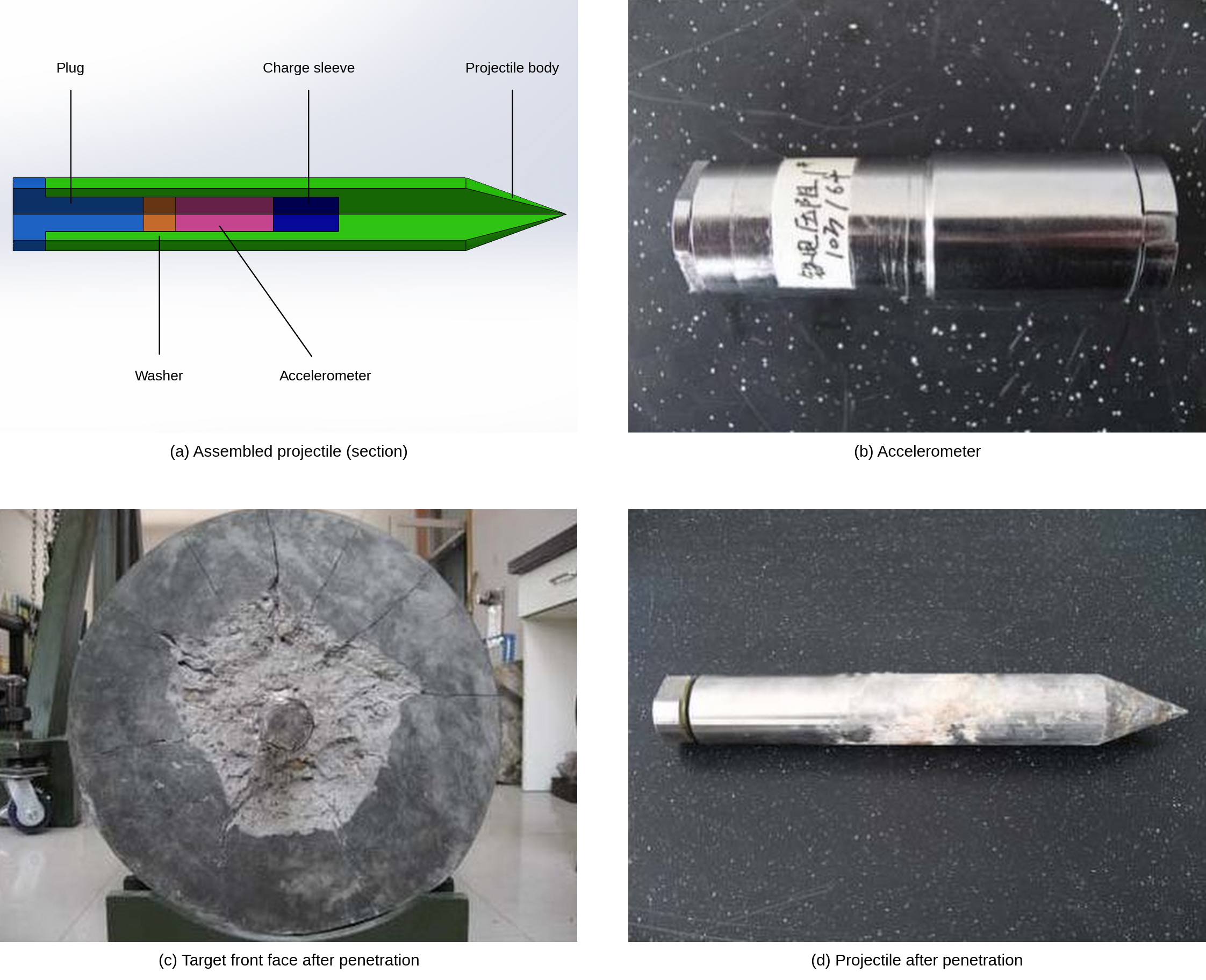}
	\caption{Projectile assembly and post-penetration morphology. (a) The assembly section of the projectile; (b) the acceleration sensor; (c) the front target face after penetration; (d) the projectile after penetration.}
	\label{fig:4}
\end{figure}

The assembly section of the projectile, the acceleration sensor, the front target face after penetration and the projectile after penetration in the second group of tests are shown in Figure~\ref{fig:4}, where the assembly section marks the relative positions of the projectile, the charge sleeve, the washer and the acceleration sensor.

\subsection{Deceleration time history and verification of the numerical model}
\label{sec:3_2}

\begin{figure}[!ht]
	\centering
	\includegraphics[width=\linewidth]{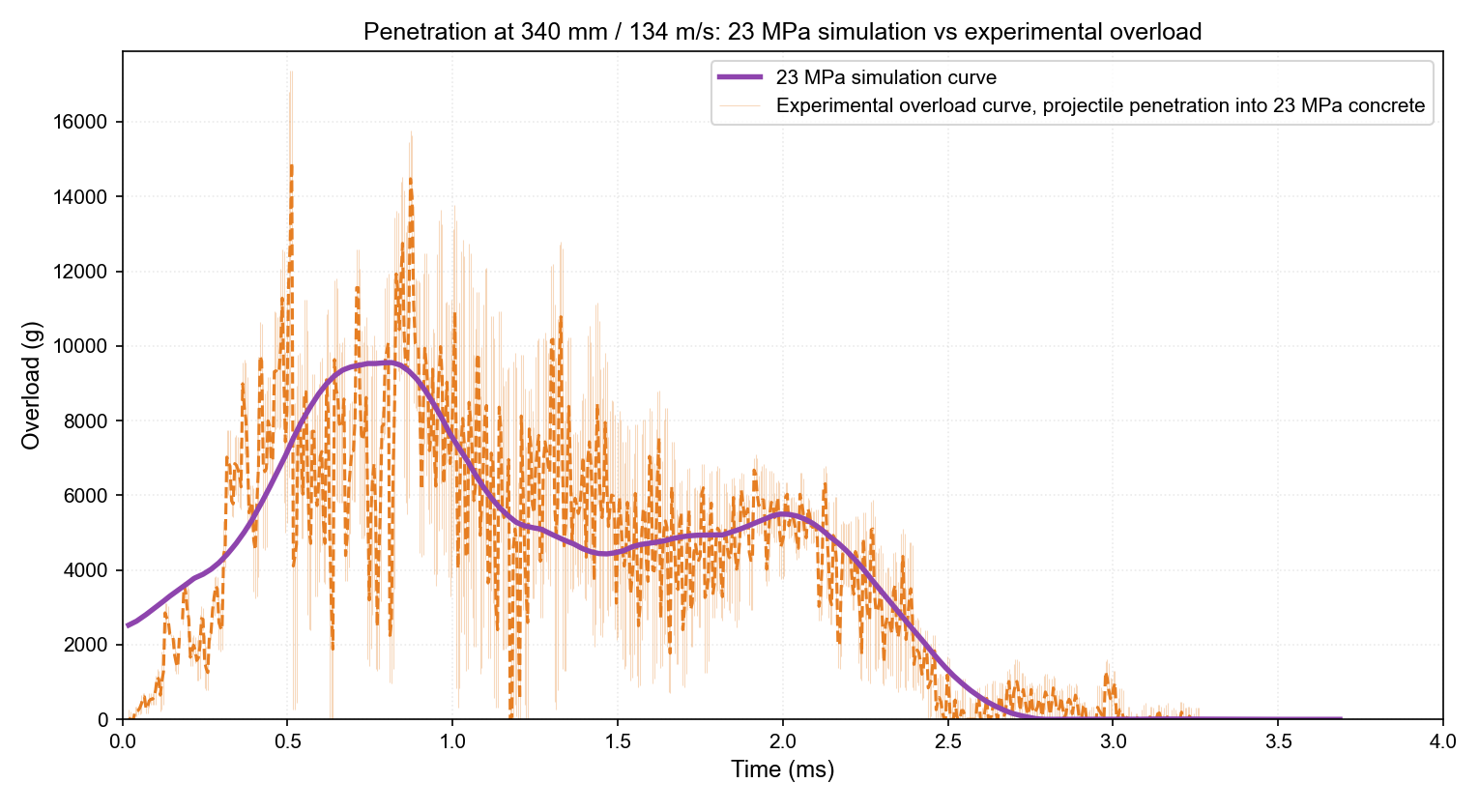}
	\caption{Comparison between the simulated and the measured deceleration time histories. The simulation is the case with a total projectile length of 340 mm, an impact velocity of 134 m/s and a target plate compressive strength of 23 MPa; the measurement is the deceleration curve recorded in the second group of tests on the same apparatus and at the same target strength. The horizontal axis is time and the vertical axis is the projectile deceleration.}
	\label{fig:5}
\end{figure}

The deceleration time history measured in the second group of tests is given by the measured branch in Figure~\ref{fig:5}; the duration of the deceleration is 2.51 ms, and the curve exhibits dense high-frequency oscillations near 0.5 ms caused by the axial vibration of the projectile.

The credibility of numerical simulations of structural impact is mostly verified by comparison with tests or by international collaborative benchmarks \cite{ref37}. A case with a 23 MPa target plate was established with the material models, the constitutive models and the discretization of Chapter~\ref{sec:2}; the total projectile length was taken as the value of the second group of tests, 340 mm, and the impact velocity as 134 m/s. The resulting deceleration time history is compared with the measured curve in Figure~\ref{fig:5}. The deceleration time history given by the simulation and the measured oscillatory curve are consistent in magnitude and duration, and both fall to zero near 2.5 to 2.6 ms. The experimental projectile and the simulated projectile are identical in total length and impact velocity and differ in the nose cone angle and the cavity dimensions, this difference being constrained by the experimental conditions; the comparison in Figure~\ref{fig:5} is therefore limited to the magnitude and the duration under the same apparatus, the same velocity level and the same target strength.

The above results indicate that the mesoscale modelling approach of Chapter~\ref{sec:2} is able to reproduce the deceleration magnitude and the deceleration duration of the laboratory ballistic penetration tests, so that the stress field generated by that approach can be regarded as a credible data source for the training of the field-level operator in Chapter~\ref{sec:5}. The object verified in this chapter is the explicit finite element simulation and the data source generated by it, and the accuracy of the field-level operator itself is given in Chapter~\ref{sec:6}.

\subsection{Crater morphology}
\label{sec:3_3}

\begin{figure}[!ht]
	\centering
	\includegraphics[width=\linewidth]{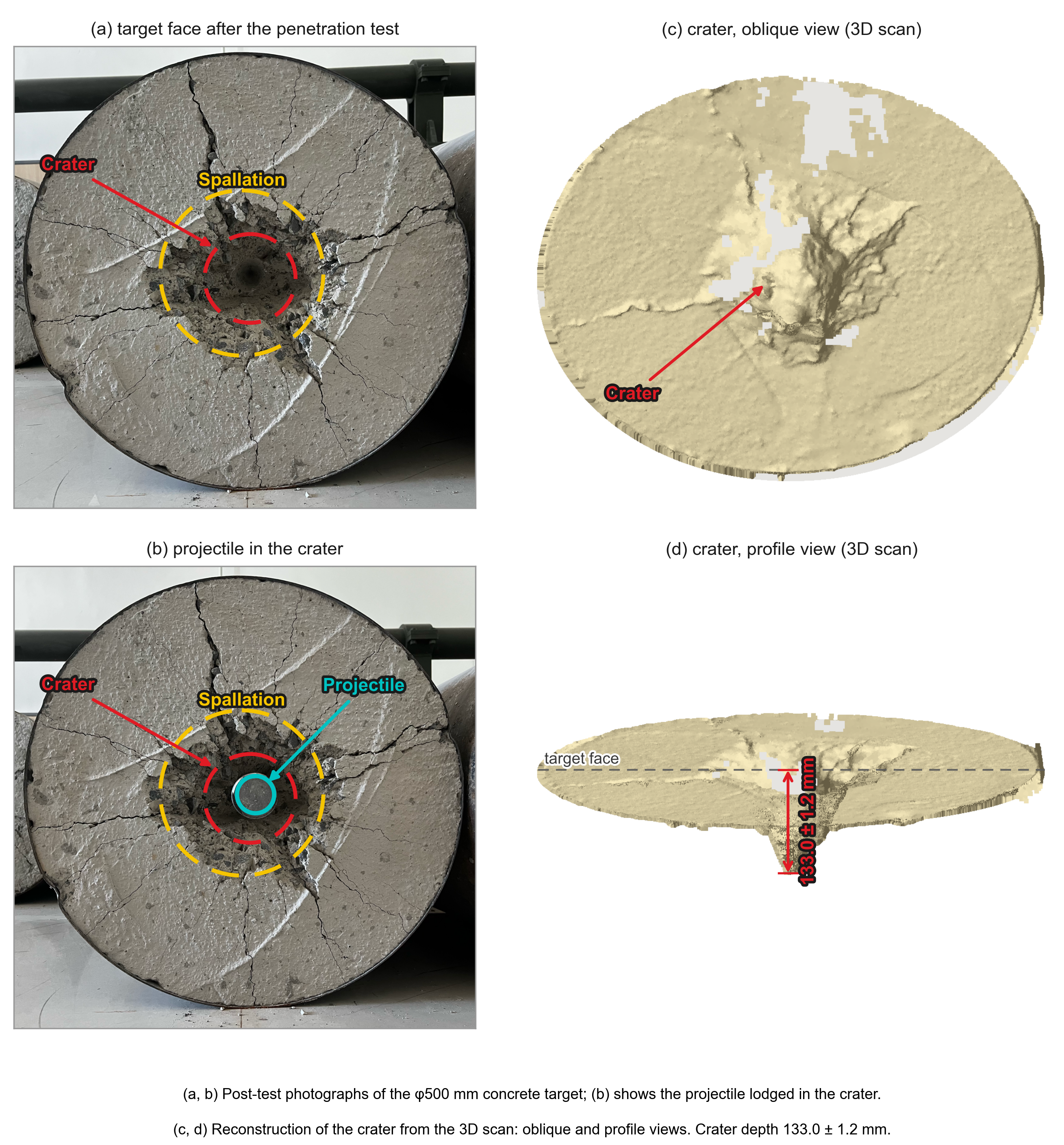}
	\caption{Crater morphology and section. (a) The target face after penetration, with the crater and the annular spalling zone marked; (b) the projectile embedded in the crater; (c) the three-dimensional laser scanning reconstruction of the crater; (d) the section profile through the crater centre, with a crater depth of 133.0 $\pm$ 1.2 mm.}
	\label{fig:6}
\end{figure}

The morphology of the target face after the first group of tests was recorded with a camera, the crater was reconstructed by three-dimensional laser scanning, and the crater depth was measured with a depth probe. The crater morphology is shown in Figure~\ref{fig:6}. An annular spalling zone is formed on the target face around the projectile hole, the internal profile of the crater is irregular, and local spalling is present on the crater wall. The crater depth measured from the section is 133.0 $\pm$ 1.2 mm. After the test the projectile exhibited abrasion only at the nose, and no visible deformation occurred on the projectile body, so that the rigid-body projectile assumption adopted in Chapter~\ref{sec:2} is consistent with the experimental observation.

\section{Dataset}
\label{sec:4}

\subsection{Composition of the cases}
\label{sec:4_1}

\begin{figure}[!ht]
	\centering
	\includegraphics[width=\linewidth]{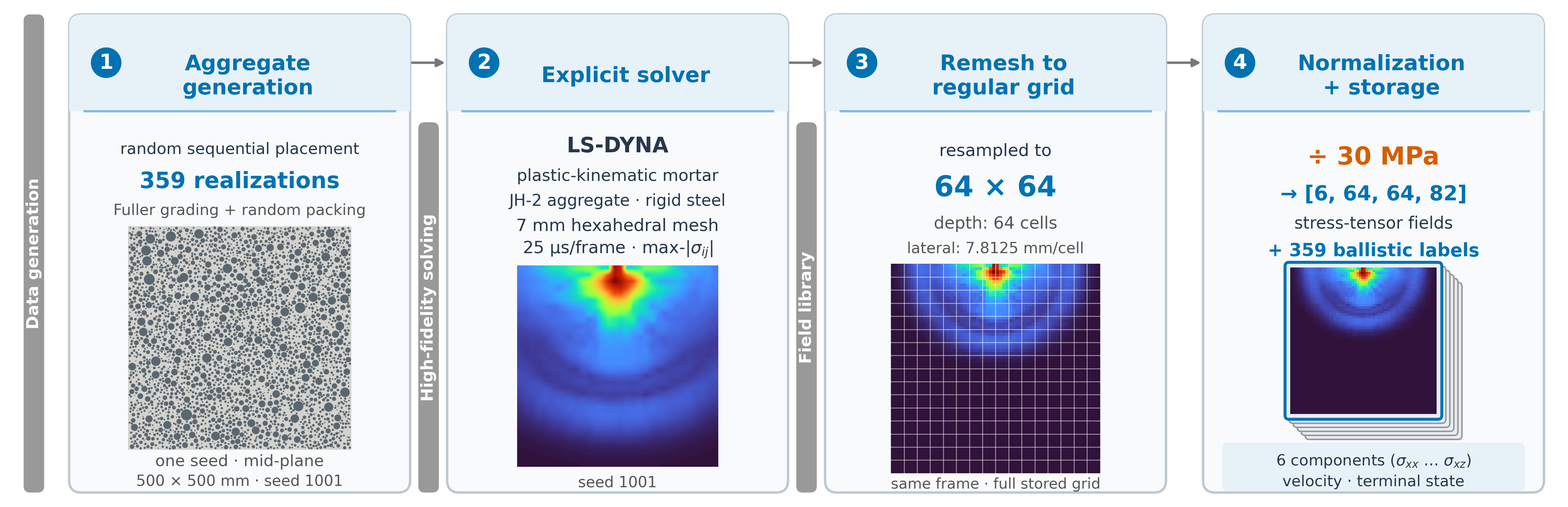}
	\caption{Dataset generation workflow. Four steps in total, which are aggregate generation (parameter setting and random placement), the explicit solution, the resampling to a regular grid and normalization and storage, corresponding in order to the numbered panels 1 to 4 of the figure. The stress tensor field is stored as a four-dimensional array of $[6,64,64,82]$, and the field quantities are linearly normalized in the export layer.}
	\label{fig:7}
\end{figure}

The dataset was generated with the mesoscale numerical model of Chapter~\ref{sec:2} and comprises 359 cases, which are divided into three levels by impact velocity. The impact velocities of the three levels are 100, 135 and 165 m/s, the corresponding concrete compressive strengths are 14, 23 and 48 MPa, and the numbers of cases are 149, 107 and 103 respectively. Within one level the cases differ only in the aggregate random seed, while the material parameters and the solution settings are completely identical, so that these cases are geometrically independent of each other and physically identically distributed; between different levels the impact velocity and the concrete compressive strength vary simultaneously. The cases of each level that take part in training and evaluation are first subjected to one shuffled permutation of the whole set with a fixed random seed and are then divided into the three parts of training, validation and test in the ratio $7:1.5:1.5$; the random seed used for the division is independent of the seed used for the model initialization and takes the same value for all the methods compared, so that different methods are compared on one and the same division. The aggregate seeds of the test part do not appear in the training and validation parts, and the training and validation parts are used for the fitting and the selection of the operator, the test part taking no part in the selection. The composition of the dataset is given in Table~\ref{tab:2} and its generation workflow in Figure~\ref{fig:7}.

\begin{table}[!ht]
\centering
\caption{Composition of the dataset}\label{tab:2}
\footnotesize
\setlength{\tabcolsep}{2pt}
\begin{tabular}{l >{\raggedright\arraybackslash}p{1.8cm} >{\raggedright\arraybackslash}p{1.7cm} c c c c c}
\toprule
Level & \makecell[t]{Concrete\\compressive\\strength\\/ MPa} & \makecell[t]{Impact\\velocity\\/ (m$\cdot$s$^{-1}$)} & Cases & \makecell[t]{Training\\cases} & \makecell[t]{Validation\\cases} & \makecell[t]{Test\\cases} & \makecell[t]{Evaluation\\windows} \\
\midrule
v100 & 14 & 100 & 149 & 83 & 17 & 19 & 608 \\
v135 & 23 & 135 & 107 & 60 & 12 & 14 & 476 \\
v165 & 48 & 165 & 103 & 58 & 12 & 14 & 476 \\
\midrule
Total & --- & --- & \textbf{359} & \textbf{201} & \textbf{41} & \textbf{47} & \textbf{1560} \\
\bottomrule
\end{tabular}

\vspace{2pt}
\begin{minipage}{\textwidth}
\footnotesize
Note: the number of cases is the total number of cases at that level; the numbers of cases in the training, validation and test parts are obtained by dividing the cases of that level that take part in training and evaluation in the ratio $7:1.5:1.5$, the numbers of cases taking part in the division being 119, 86 and 84 for the three levels respectively, and each level additionally contains cases that are not used in the present work; the aggregate seeds of the test cases do not appear in the training and validation parts. An evaluation window denotes a single-step prediction sample taken from the test cases under the setting of Chapter~\ref{sec:6}; the number of windows is determined jointly by the number of multi-step rollout steps adopted in the training of that level and by the starting-point protocol, and is 608, 476 and 476 for the three levels respectively, which corresponds to the range of target frames covered by the evaluation of each level, the ranges being given in the note to Table~\ref{tab:3} of Chapter~\ref{sec:6}; a comparison across levels must report the number of windows together, since comparability holds only within one and the same level.
\end{minipage}
\end{table}

\subsection{Field quantities}
\label{sec:4_2}

The mid-plane stress field of each case is obtained by resampling the element stresses output by the explicit solution and is a four-dimensional array of shape $[6,64,64,82]$, the component order and the grid convention being given in Section~\ref{sec:2_4}. In addition to the six-component stress field, three groups of auxiliary fields are stored on the same cross-section: the velocity field and the displacement field each contain three components and have the shape $[3,64,64,82]$; the equivalent plastic strain and the compressive damage and the tensile damage in the z direction are single-component fields of shape $[64,64,82]$. The aggregate volume fraction field is stored separately as a three-dimensional array of shape $[64,64,64]$, whose third dimension is the out-of-plane direction. All field quantities share the same time discretization, namely 82 frames with a frame interval of 0.025 ms.

\subsection{Ballistic labels}
\label{sec:4_3}

\begin{figure}[!ht]
	\centering
	\includegraphics[width=\linewidth]{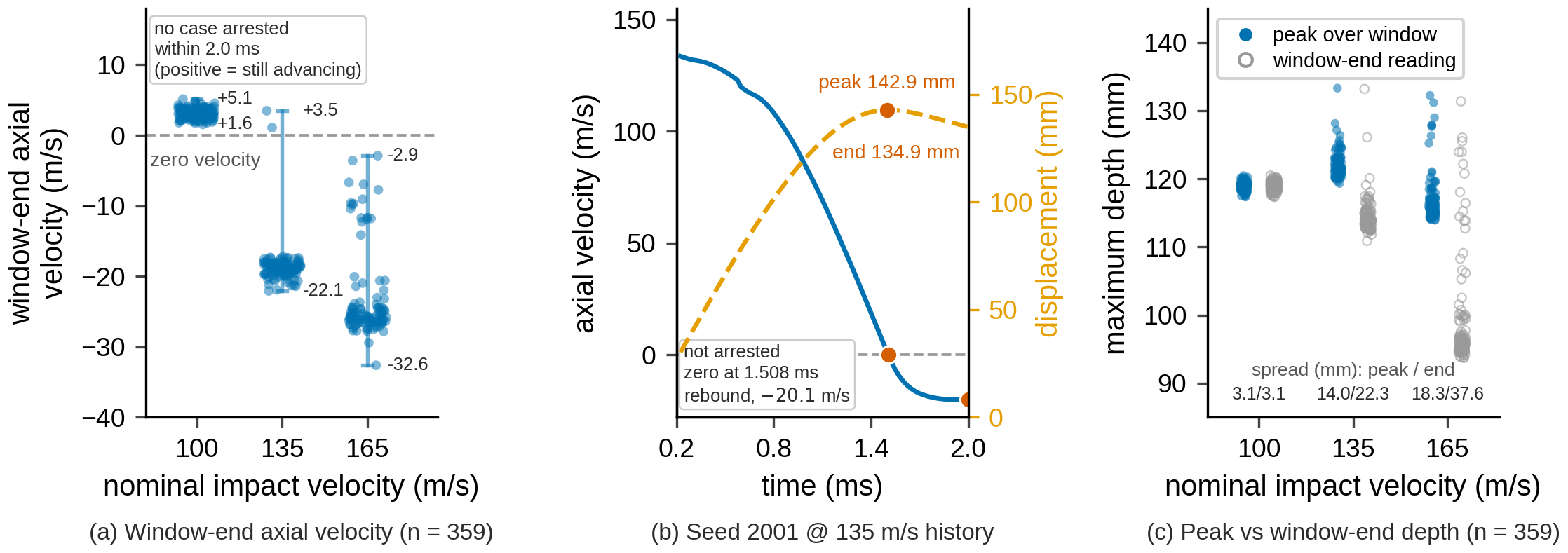}
	\caption{Terminal-state verification of the three velocity levels. (a) The distribution over cases of the projectile axial velocity at the end of the window, a positive value meaning that the projectile is still advancing into the target; (b) the velocity and displacement time histories of one case at the 135 m/s level; (c) the scatter over cases of the peak penetration depth within the window and the penetration depth at the end of the window. The numbers of cases of the three levels are 149, 107 and 103 respectively.}
	\label{fig:8}
\end{figure}

Each case is accompanied by a ballistic label table containing 82 rows, one row per frame, with the columns of frame index, time, absolute depth of the projectile tip, projectile displacement, projectile velocity, z-direction contact force on the projectile, system kinetic energy and solution state. The projectile deceleration is obtained by dividing the contact force by the projectile mass and is used for the curve verification of Chapter~\ref{sec:3} and for evaluation at the level of engineering quantities. Each case is further accompanied by a metadata file recording the projectile and target geometry, the material parameters, the mesh size and the grid convention of the field quantities, which is provided together with the field files and with this table. The projectile axial velocity and the projectile tip depth of the three velocity levels at the termination time of the solver, 2.000 ms, are shown in Figure~\ref{fig:8}.

\subsection{Data and code availability}
\label{sec:4_4}

The stress field files, the auxiliary field files, the ballistic label tables and the metadata files of the dataset used in the present work, together with the implementation, training and evaluation programs of the operator, can be obtained from the corresponding author upon reasonable request. The dataset was generated with the mesoscale numerical model of Chapter~\ref{sec:2}, and the reproducibility claim of the present work rests on the field data, the ballistic labels and the operator programs described above.

\section{Operator}
\label{sec:5}

\begin{figure}[!ht]
	\centering
	\includegraphics[width=\linewidth]{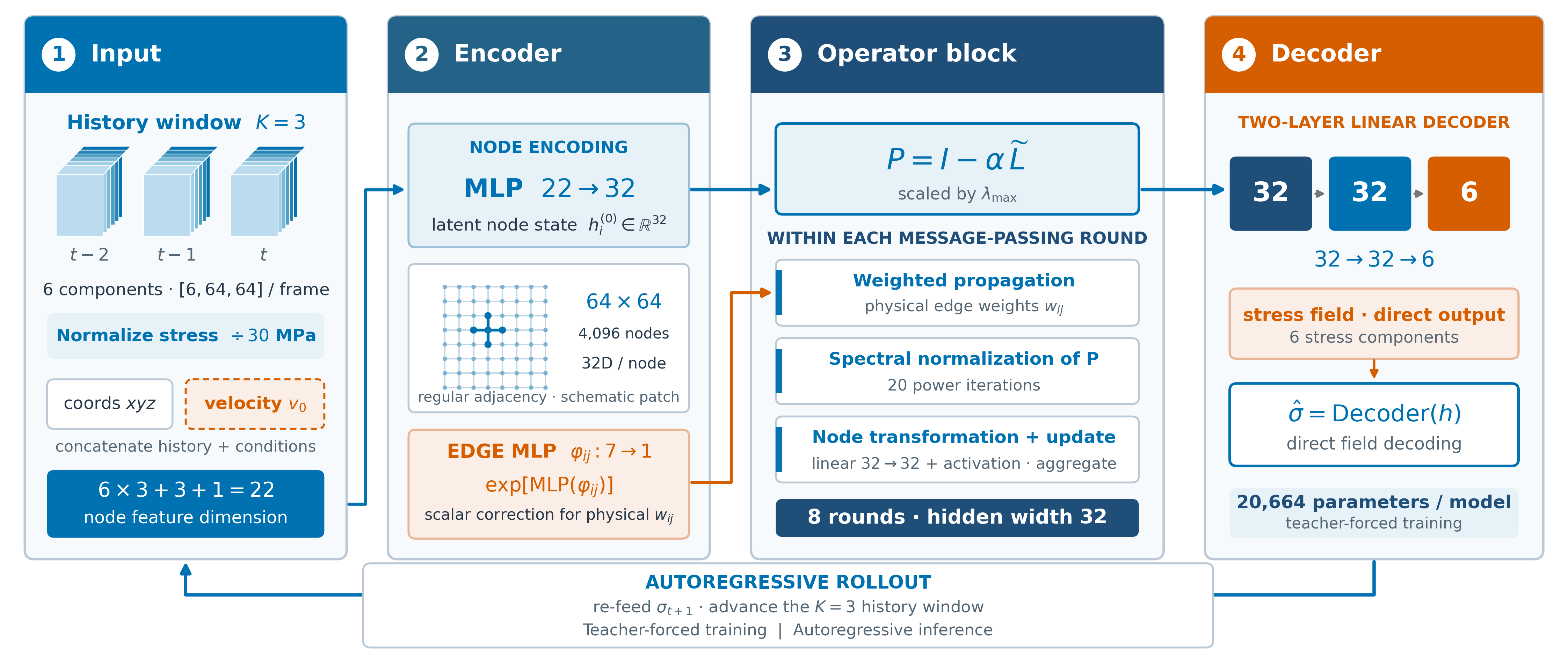}
	\caption{Overall structure of the field-level neural operator. The input is the normalized six-component stress tensor field of consecutive frames within the look-back window together with the impact velocity scalar, and the encoder concatenates the nodal stress components, the nodal coordinates and the impact velocity into node features and compresses them into a latent representation; the operator layers perform 8 rounds of message passing on the cross-section node graph with a latent width of 32, the aggregation being weighted by edge weights constructed from physical quantities, and the propagation operator applying a graph Laplacian spectral normalization (20 power iterations); the decoder outputs the six stress components of the next frame through two linear mappings. The number of nodes shown in the figure is $64\times64$ of the storage grid; the grid on which the operator of the present work actually operates is $32\times32$, that is, 1024 nodes, and the relation between the two is given in Section~\ref{sec:5_2}.}
	\label{fig:9}
\end{figure}

This chapter presents the construction of the mid-plane stress field operator. The input of the operator is the stress field of the most recent frames of one and the same case together with the projectile impact velocity, and the output is the stress field of the next frame; the spatial variation of the aggregate arrangement enters the operator directly without any low-dimensional compression. The problem formulation, the network architecture, the construction of the edge weights, the propagation operator and the training configuration are presented in order, and the overall structure of the operator is shown in Figure~\ref{fig:9}.

\subsection{Problem formulation}
\label{sec:5_1}

The mid-plane stress field is denoted by $\sigma_t \in \mathbb{R}^{6\times32\times32}$, where $t$ is the frame index, $X$ is the spatial support formed by the $32\times32$ regular node set obtained by sampling every other point on the mid-plane, and $v_0$ is the impact velocity of the case. The single-step form of the operator is taken as

\begin{equation}
\label{eq:1}
\sigma_{b_0+16} = \mathcal{G}_{\theta}\left(\sigma_{b_0-32},\ \sigma_{b_0-16},\ \sigma_{b_0};\ X,\ v_0\right)
\end{equation}

The input consists of three historical stress fields with a frame interval of 16 frames, which is identical to the prediction span, so that a single call advances the solution by 0.4 ms and the three historical frames cover 0.8 ms of evolution information. The impact velocity enters the operator as a global condition.

Both the input and the output of the mapping are field quantities defined on the same spatial grid, and the spatial structure of the field is not compressed by a low-dimensional parameterization. The local stress concentrations, the wave front shape and the damage distribution caused by the difference in the aggregate arrangement therefore enter the operator in the form of field quantities, without low-dimensional parameterization compression; what the operator learns is thus not a regression from a low-dimensional feature vector to the field, but an extrapolation from field to field. The three historical frames are taken at equal intervals, so that the operator sees a series of states at equal time steps rather than a small increment between adjacent frames.

\subsection{Network architecture}
\label{sec:5_2}

The mid-plane field is taken as a graph: every other point on the mid-plane grid obtained by the resampling of Section~\ref{sec:2_4} becomes a node, giving 1024 nodes in total with a node spacing of 15.625 mm; the nodes are connected by edges according to the grid adjacency, each node being connected to its four face-adjacent nodes. Both the node set and the adjacency are determined by this regular sampling, and no additional graph construction hyperparameter is introduced.

The dimension of the node feature is 22. The six stress components of each of the three historical frames constitute 18 dimensions, the coordinates of the node on the cross-section constitute 3 dimensions, and the impact velocity of the case constitutes 1 dimension. The three kinds of information characterize respectively the local stress state at the node, the position of the node on the cross-section and the impact condition of the case. The 22-dimensional feature is encoded into a 32-dimensional latent state by a two-layer perceptron with layer normalization, and the components are scaled by the normalization coefficients of Section~\ref{sec:2_4}.

Message passing comprises 8 rounds with a latent width of 32 and adopts the form common in graph neural networks \cite{ref38}. The update of each round is composed of three terms: the neighbourhood latent states are aggregated along the graph edges, the aggregation strength being given by the graph Laplacian propagation operator; a node-wise residual transformation is added; and the initial latent state is mixed in at a fixed ratio, after which layer normalization is applied. The construction of the propagation operator is given in Section~\ref{sec:5_4} and that of the edge weights in Section~\ref{sec:5_3}.

The decoder consists of two linear mappings and maps the 32-dimensional latent state to the six stress components at the nodes, the output being the mid-plane stress field of the next frame. The decoding is performed independently at the nodes, and the spatial organization of the output field is determined entirely by the message passing stage, so that both the smooth segments of the wave front and the high-gradient regions near the aggregates result from the propagation of information on the graph.

The total number of parameters of the operator is 20\,664. The parameters appear only in the encoder, the per-round transformations, the edge weight correction term and the decoder, and do not grow with the number of cases.

\subsection{Physically coupled edge weights}
\label{sec:5_3}

The strength of message passing along a graph edge is determined by the edge weight $w_{ij}$. This edge weight is given by the product of three factors:

\begin{equation}
\label{eq:2}
w_{ij} = T_{ij} \cdot \|t_i\| \cdot \exp\left(\mathrm{clamp}\left(\mathrm{MLP}(\varphi_{ij}),-3,3\right)\right)
\end{equation}

in which the first factor $T_{ij}$ is the harmonic mean of the acoustic impedances on the two sides of the interface,

\begin{equation}
\label{eq:3}
T_{ij} = \frac{2 Z_i Z_j}{Z_i + Z_j}
\end{equation}

The nodal acoustic impedance $Z$ is given by the aggregate volume fraction at that node according to the Voigt upper bound of a two-phase composite: the P-wave modulus is first obtained from the elastic modulus and the Poisson ratio, the acoustic impedance is then obtained from the P-wave modulus and the density, and the result is normalized to the matrix phase. The second factor $\|t_i\|$ is the magnitude of the Cauchy traction of node $i$ on the corresponding face, that is, the square root of the sum of the squares of the normal stress and the shear stress, so that the normal component and the shear stress both enter the transfer strength. The third factor is a data-driven correction term whose input $\varphi_{ij}$ is a 7-dimensional edge feature, formed by the aggregate volume fractions at the two end nodes, the stress norms at the two end nodes, the signed normal projection of the traction and the values of the impact velocity at the two end nodes.

The three factors play different roles. The first factor brings the impedance contrast on the two sides of the interface directly into the information transfer strength, so that the material heterogeneity takes part in the construction of the transfer strength; the second factor makes a node at a higher local stress level transfer more information to its neighbourhood, and the transfer direction changes as the wave front advances; the third factor carries the data-driven correction, and its output layer is zero-initialized so that $\exp(0) = 1$, with the edge weight at the beginning of training being determined entirely by the first two physical factors. The signed normal projection enters the correction term as a feature, so that the correction under the two stress states of compression and tension can be learned separately. The edge weight is symmetrized and then truncated to $[1\times10^{-6}, 100]$.

The construction of the edge weight gives one and the same set of propagation rules the same physical meaning on cases with different aggregate arrangements and at different velocity levels: the impedance contrast comes from the local volume fraction, the traction magnitude comes from the local stress state, and the impact velocity enters the correction term in the form of a conditional feature, so that no interface parameter needs to be specified separately between cases.

\subsection{The spectrally normalized propagation operator}
\label{sec:5_4}

The message passing is defined on the normalized graph Laplacian \cite{ref39}. Let $A$ be the adjacency matrix formed by the symmetrized edge weights, $D$ the corresponding degree matrix, $L = D - A$ the graph Laplacian matrix, and $\lambda_{\max}$ the largest eigenvalue of this matrix, which is estimated for the graph by 20 power iterations at each forward propagation. The propagation operator is taken as

\begin{equation}
\label{eq:4}
P = I - \alpha \tilde{L}, \qquad \tilde{L} = L / \lambda_{\max}
\end{equation}

in which $\alpha$ is a learnable scalar parameterized by the sigmoid function, whose value lies in the interval $[0, 1]$.

The eigenvalues of $\tilde{L}$ lie in the interval $[0, 1]$, so that the eigenvalues of $P$ lie in the interval $[1 - \alpha, 1]$. The normalization factor is given by the power iteration estimate, and this restriction on the spectral radius holds in the estimation sense rather than as a strict structural guarantee.

\subsection{Training objective and configuration}
\label{sec:5_5}

The training target is the absolute stress field, that is, the operator directly outputs the six stress components of the next frame rather than the increment relative to the current frame. The loss is taken as the weighted sum of the relative $L_2$ error of each step in the multi-step rollout under teacher forcing, with weights decreasing along the rollout steps; the historical frames used in teacher forcing are taken from the solver ground truth rather than from the output of the model itself. For a three-step rollout the weights of the steps are taken as 0.5, 0.3 and 0.2 respectively, so that the prediction closer to the current frame carries a higher weight in the loss; for a single-step rollout this term degenerates into the single-step relative $L_2$ error.

The relative $L_2$ error is computed only on the active nodes. An active node is defined as a node at which the amplitude of the ground-truth stress field exceeds a given threshold, the threshold being taken as 0.01 after normalization, which corresponds to 0.3 MPa under the normalization coefficient of Section~\ref{sec:4_2}. The ground-truth field in the quiescent region is close to zero, and if it were included, the error would be diluted by the signal-free region.

The primary setting is: 3 look-back frames, a historical frame interval of 16 frames, a latent width of 32 and 8 propagation rounds. This set of settings is identical across the three levels. The optimizer is AdamW with a learning rate of $3\times10^{-3}$.

The training budget is set by the maximum number of epochs of each level: the 100 m/s level is trained with a three-step rollout and a maximum of 60 epochs; the 135 and 165 m/s levels are trained with a single-step rollout and a maximum of 200 epochs. One and the same early stopping criterion is adopted for the three levels, namely termination within the maximum number of epochs when the validation error no longer decreases, so that the actual training end point is determined by this criterion and the maximum number of epochs is only an upper bound.

Model selection is based on the weighted relative $L_2$ error on the validation set. The validation set consists of 50 fixed windows; during training, a parameter snapshot is saved whenever this error reaches a new minimum, and training stops when no new minimum is reached for 40 consecutive epochs, the snapshot with the lowest validation error being used for the evaluation on the test part; the test part plays no part in the termination criterion of the training or in the parameter selection.

\section{Results}
\label{sec:6}

This chapter presents the field-level accuracy of the physics-coupled graph neural operator (PC-GNO). The evaluation protocol, the results at the three velocity levels and the comparison with the baselines under the same setting are presented in order.

\subsection{Evaluation protocol}
\label{sec:6_1}

The accuracy is defined by the relative $L_2$ error, and the accuracy percentage is computed as 1 minus the relative $L_2$ error, multiplied by 100, so that a larger value is better. The relative $L_2$ error is computed frame by frame as $\|p - g\| / \|g\|$ and then averaged across the windows, where $p$ is the predicted field of the operator and $g$ is the solver ground-truth field of the same window. The error is computed only on the active nodes, and both the numerator and the denominator take the norm on the active node set, that is, $\|p - g\|$ and $\|g\|$ are both evaluated only on the active nodes, so that the denominator is not the ground-truth norm over the whole cross-section; the determination of the active nodes is identical to that used in the training loss of Section~\ref{sec:5_5}. The per-window error is normalized by the ground-truth norm of that window, and the windows are then averaged with equal weight.

The starting protocol is the single-step form, the historical frames being taken as the three frames $b_0 - 32$, $b_0 - 16$ and $b_0$ and the prediction target being frame $b_0 + 16$, in accordance with Eq.~(\ref{eq:1}) of Chapter~\ref{sec:5}. Accordingly, the target frames covered by the evaluation start from frame 48, extending to frame 79 at the 100 m/s level and to frame 81 at the 135 and 165 m/s levels; the numbers of windows are correspondingly 608, 476 and 476. The windows are taken as all the available windows of this span of the test cases of each level, with equal weight per window, and no window is selected and no per-frame optimum is taken. The aggregate seeds of the test cases do not appear in the training or the validation part.

\subsection{Field-level accuracy at the three velocity levels}
\label{sec:6_2}

The field-level accuracy of the three levels is listed in Table~\ref{tab:3}. The average accuracy of the three levels is \textbf{80.16\%}; the 100, 135 and 165 m/s levels give \textbf{91.45\%}, \textbf{77.80\%} and \textbf{71.22\%} respectively, and the gains relative to the preceding-frame field of the same window are \textbf{27.81}, \textbf{42.46} and \textbf{34.16} percentage points respectively. The gain has the same sign at the three levels, each of them being higher than the preceding-frame field of the same window.

\begin{table}[!ht]
\centering
\caption{Field-level accuracy at the three velocity levels}\label{tab:3}
\footnotesize
\setlength{\tabcolsep}{3pt}
\begin{tabular}{>{\raggedright\arraybackslash}p{2.6cm} >{\raggedright\arraybackslash}p{1.8cm} c c c c}
\toprule
Velocity level & \makecell[t]{Concrete\\compressive\\strength\\/ MPa} & \makecell[t]{Evaluation\\windows} & Accuracy \% & \makecell[t]{Preceding-\\frame\\field \%} & \makecell[t]{Gain\\(per-\\centage\\points)} \\
\midrule
100 m/s & 14 & 608 & \textbf{91.45} & 63.64 & \textbf{+27.81} \\
135 m/s & 23 & 476 & \textbf{77.80} & 35.34 & \textbf{+42.46} \\
165 m/s & 48 & 476 & \textbf{71.22} & 37.06 & \textbf{+34.16} \\
\midrule
\textbf{Average of the three levels} & --- & --- & \textbf{80.16} & --- & --- \\
\bottomrule
\end{tabular}

\vspace{2pt}
\begin{minipage}{\textwidth}
\footnotesize
Note: the reference item is the preceding-frame field. The evaluation protocol is identical across the three velocity levels: the starting protocol is single-step, the historical frame interval is 16 frames, the error is computed frame by frame and then averaged across the windows, the denominator is the ground-truth field norm of the same window, and the parameter count of the operator is 20\,664. The active nodes are determined by a threshold on the ground-truth amplitude and are independent of the prediction, so that the operator and the reference item use one and the same mask on one and the same set of windows, and the comparison is carried out within the same evaluation domain. The models of the three velocity levels are trained independently, and the target frame spans covered by the evaluation differ from level to level (frames 48 to 79 at the 100 m/s level and frames 48 to 81 at the 135 and 165 m/s levels). The window count is a per-level quantity, being 608, 476 and 476 for the three levels in order, so that any comparison across levels must report the window count as well, and comparability holds only within one and the same level. The average of the three levels is the arithmetic mean of the three per-level accuracies and serves to report the three results side by side; the three levels correspond to different impact velocities and concrete compressive strengths, and the per-level values and per-level window counts are all listed in the table. This average is an aggregate of the three per-level results rather than a single overall metric, and any interpretation across levels must return to the per-level values.
\end{minipage}
\end{table}

\subsection{Comparison with the baselines under the same setting}
\label{sec:6_3}

The comparison of PC-GNO with three baselines under the same configuration is listed in Table~\ref{tab:4}. The comparison under the same setting takes the 100 m/s level as the representative velocity level, whose number of cases and number of evaluation windows are both the largest of the three levels. The baselines comprise two configurations of the Fourier neural operator and a closed-form linear regression. All the baselines take the same input, the same window set and the same starting protocol as PC-GNO, and the baselines are not tuned separately.

\begin{table}[!ht]
\centering
\caption{Comparison of field-level accuracy under the same configuration (100 m/s level, 608 windows)}\label{tab:4}
\footnotesize
\setlength{\tabcolsep}{4pt}
\begin{tabular}{>{\raggedright\arraybackslash}p{4.3cm} c c c c}
\toprule
Model & \makecell[t]{Parameter\\count} & Accuracy \% & \makecell[t]{Preceding-frame\\field \%} & \makecell[t]{Gain\\(percentage\\points)} \\
\midrule
\textbf{PC-GNO} & \textbf{20\,664} & \textbf{91.45} & 63.64 & \textbf{+27.81} \\
Fourier neural operator (configuration m32w64l2) & 16\,788\,038 & 87.9 & 63.64 & +24.27 \\
Fourier neural operator (configuration m16w64l4) & 8\,399\,686 & 82.9 & 63.64 & +19.22 \\
Closed-form linear regression & 132 & 71.0 & 63.64 & +7.39 \\
\bottomrule
\end{tabular}

\vspace{2pt}
\begin{minipage}{\textwidth}
\footnotesize
Note: the reference item is the preceding-frame field. This table belongs to the 100 m/s level only, the window set is 608, the starting protocol is single-step, and the denominator is the ground-truth field norm of the same batch of 608 windows; the gain column is computed from the difference of the unrounded relative $L_2$ errors, so that a direct subtraction from the two-decimal accuracies listed in the table may differ by 0.01 percentage point. This table reports the comparison under the same setting using the 100 m/s level as the representative velocity level; the baselines were not evaluated at the 135 and 165 m/s levels, and this table is therefore not compared under the same setting with the three-level average of Table~\ref{tab:3}. The accuracy of PC-GNO is higher than that of all the competing configurations in this comparison under the same setting; the best configuration of the Fourier neural operator gives 87.9\%, and its number of learnable parameters is about 812 times that of PC-GNO. The parameter count of the closed-form linear regression is 132 and its accuracy is 71.0\%. The closed-form linear regression takes the same 22-dimensional node features and the same six output components as PC-GNO; under the same window set and the same starting protocol of this level, the accuracy of PC-GNO is 20.45 percentage points higher than that of this linear reference.
\end{minipage}
\end{table}

\subsection{Consistency of the physical quantities}
\label{sec:6_4}

\begin{figure}[!b]
	\centering
	\includegraphics[width=\linewidth]{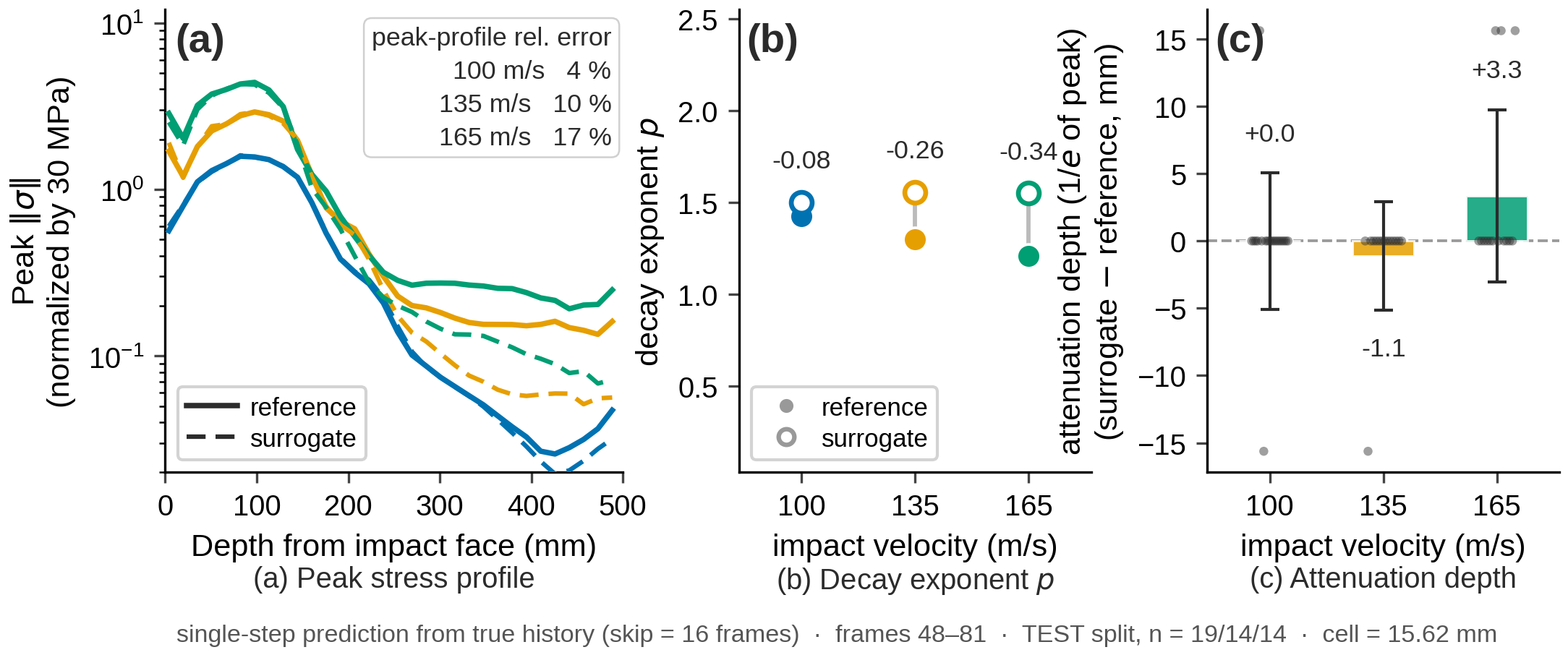}
	\caption{Peak stress profile, decay exponent and decay depth. (a) Peak stress profile along the depth direction, with a logarithmic vertical axis; (b) decay exponent of the profile; (c) decay depth difference defined by the $1/e$ drop of the peak. The reference values of the three velocity levels are all taken as the solver ground-truth field of the same window, the solid line being the reference solution and the dashed line being the operator prediction. Single-step prediction with ground-truth history and a frame interval of 16 are taken for all three levels; the evaluation windows are frames 48 to 79 (100 m/s level) and frames 48 to 81 (135 and 165 m/s levels) of the test split.}
	\label{fig:10}
\end{figure}

Besides the frame-by-frame relative $L_2$ error, the operator is also checked against several field-level physical quantities derived from the stress tensor, the check being carried out under the primary setting. Figure~\ref{fig:10} gives the peak stress profile along the depth direction, the decay exponent of this profile and the decay depth at the three velocity levels, the reference values being the solver ground-truth field of the same window.

\begin{figure}[!ht]
	\centering
	\includegraphics[width=\linewidth]{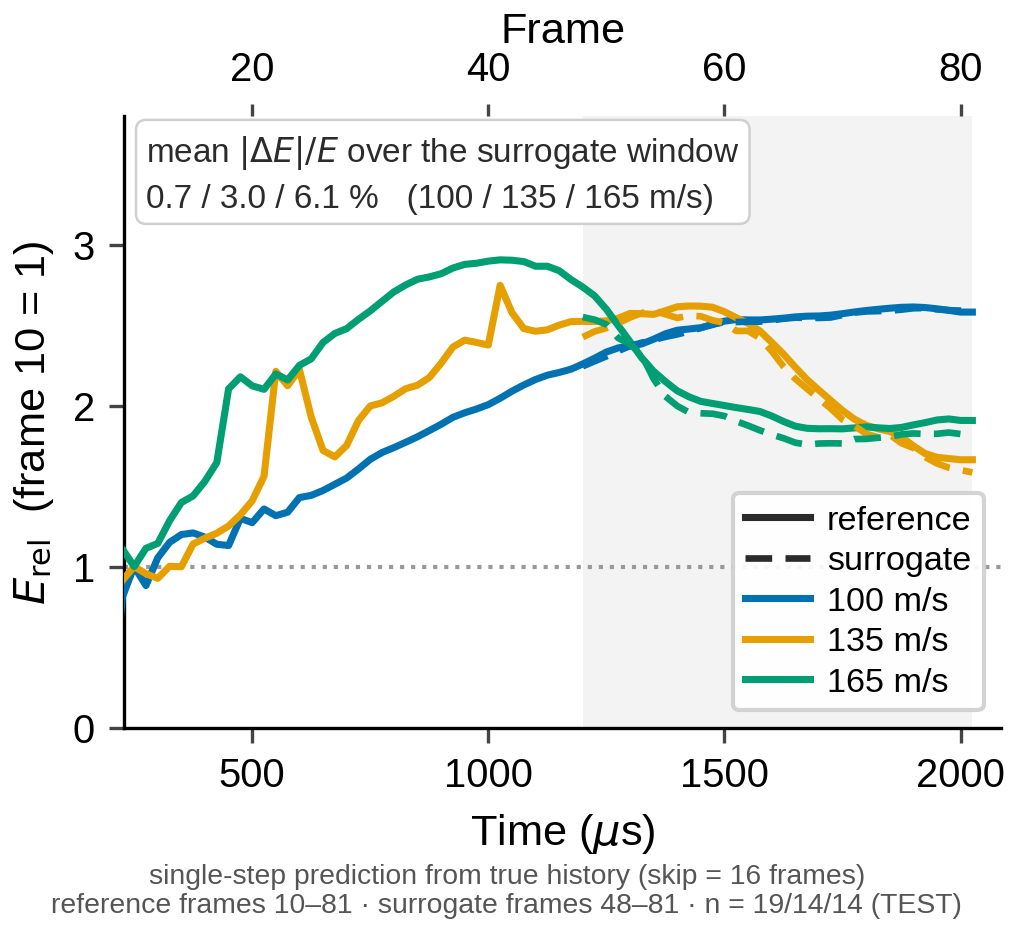}
	\caption{Evolution of the Frobenius norm of the stress tensor with time. The reference curves and the prediction curves of the three velocity levels share one and the same normalization basis, which is taken as the norm of the ground-truth field of each level at frame 10. The reference curves cover frames 10 to 81 and the prediction curves cover frames 48 to 81; the colors distinguish the impact velocity levels, the solid line being the reference solution and the dashed line being the operator prediction.}
	\label{fig:11}
\end{figure}

Figure~\ref{fig:11} gives the evolution of the Frobenius norm of the stress tensor with time; the reference curves and the prediction curves of the three velocity levels share one and the same normalization basis, which is taken as the norm of the ground-truth field of each level at frame 10. The reproduction of the above physical quantities by the operator is of the same order as the field-level accuracy of Table~\ref{tab:3}.

\subsection{Field morphology and node-level residuals}
\label{sec:6_5}

The weights used in this section are trained on the 135 m/s level by leaving one aggregate seed out, with a historical frame interval of 4 frames; the checks of the field morphology and of the node-level residuals in the following two figures hold equally under this setting, and their numbers take no part in the comparison under the same setting of Table~\ref{tab:3} and Table~\ref{tab:4}.

\begin{figure}[!ht]
	\centering
	\includegraphics[width=\linewidth]{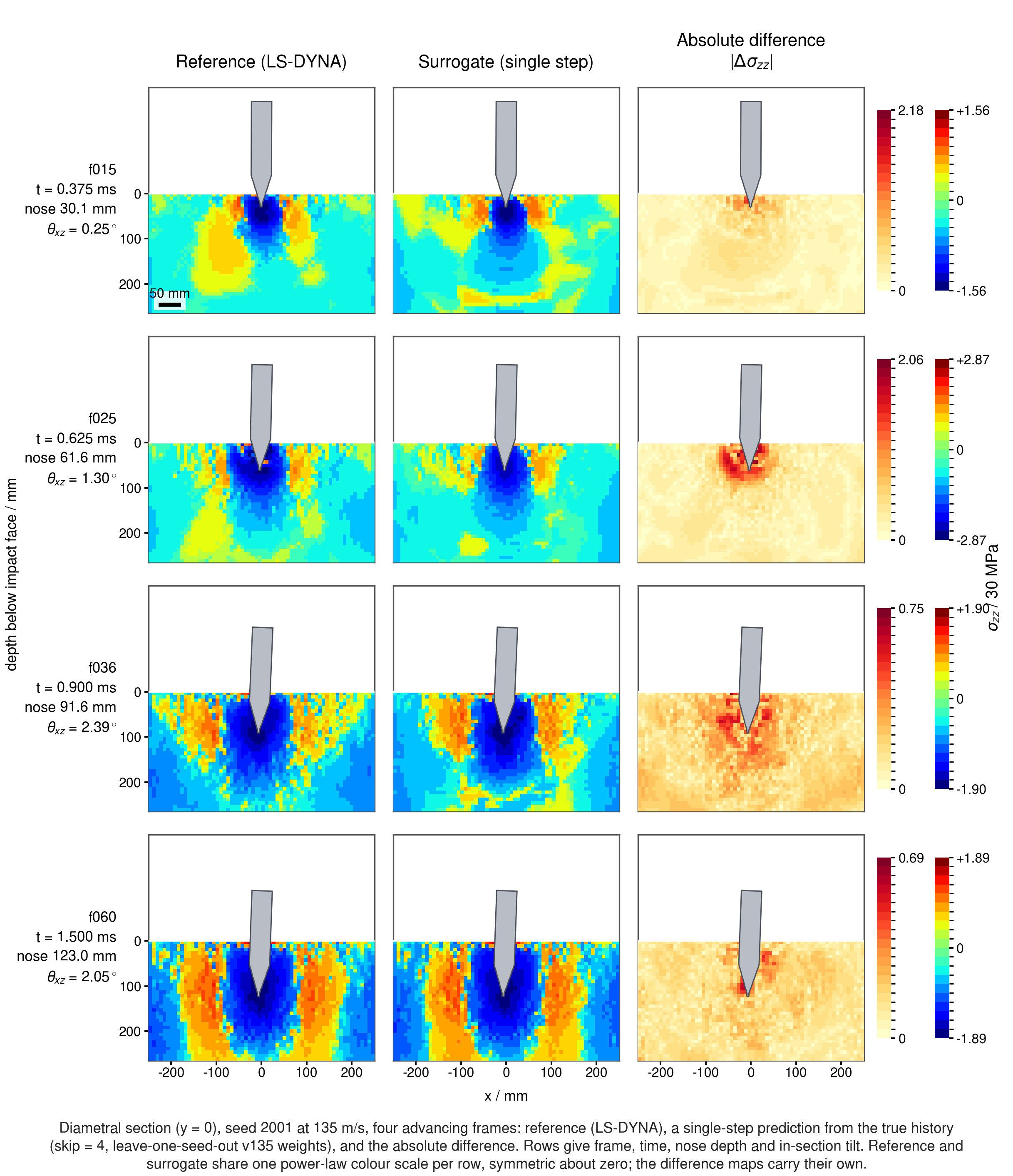}
	\caption{Comparison of the predicted field with the solver reference field. Taking one test case of the 135 m/s level as an example, the four rows give the solver reference field, the single-step prediction field of the operator and the absolute difference between the two at four advancing frames. The reference field and the prediction field of the same row share one color scale centered at zero, and the error column uses a separate color scale; the stress values are normalized values.}
	\label{fig:12}
\end{figure}

A qualitative comparison of the field morphology is shown in Figure~\ref{fig:12}. Taking one test case of the 135 m/s level as an example, the figure gives the solver reference field, the single-step prediction field of the operator and the absolute difference between the two at four advancing frames; the spatial extent of the predicted field in the compression region and its wave front position both correspond to those of the solver reference field.

\begin{figure}[!ht]
	\centering
	\includegraphics[width=\linewidth]{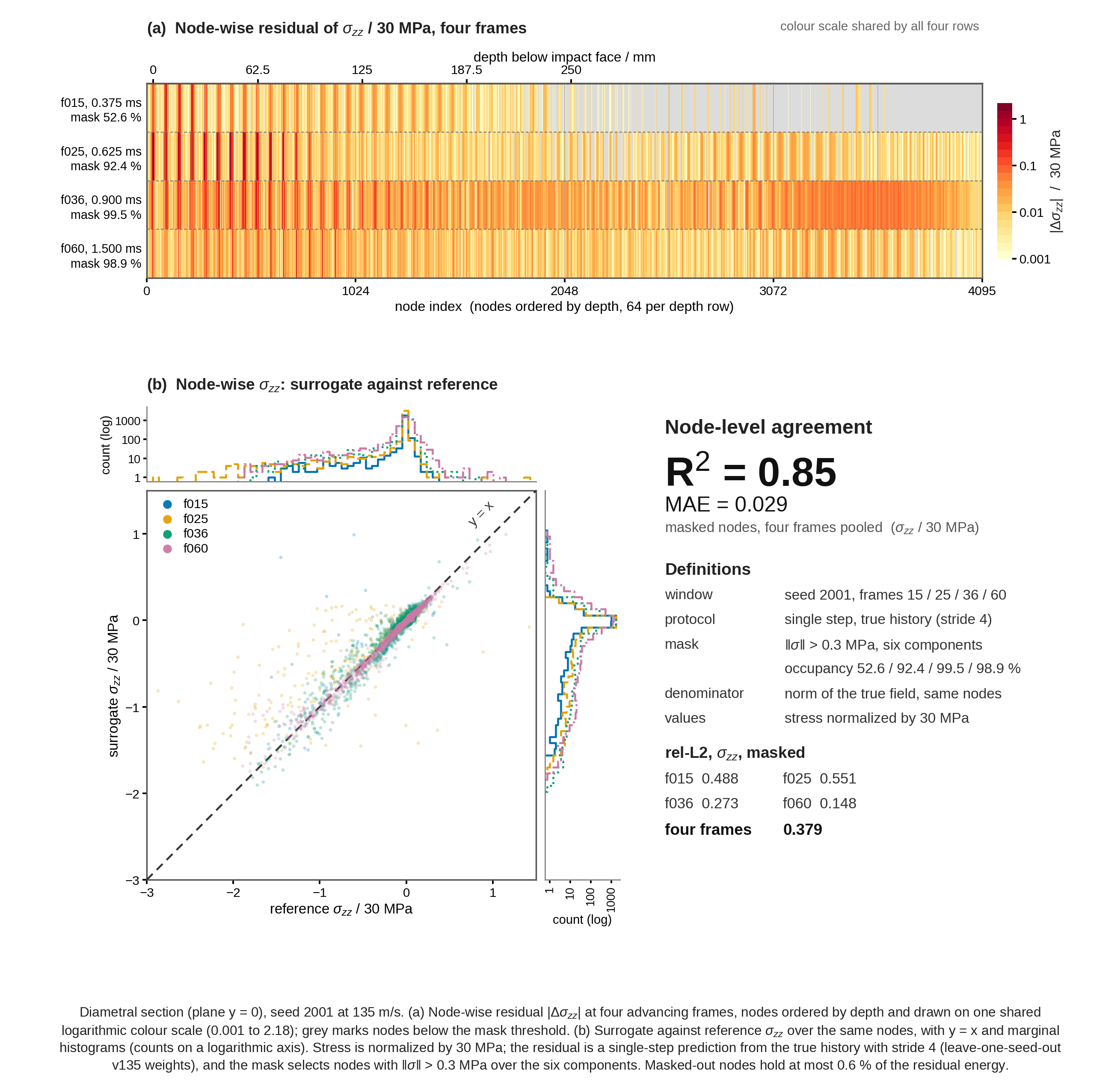}
	\caption{Distribution of the node-level residuals. (a) Node-wise $\sigma_{zz}$ absolute residual at four advancing frames, the horizontal axis being the node index ordered by depth, the error being displayed on a logarithmic color scale, and the nodes outside the mask being marked in gray; (b) node-wise comparison of the predicted values with the reference values, the diagonal being the line of equality. The node set and the conditions of evaluation of the node-level coefficient of determination differ from those of the field-level relative $L_2$ error of Table~\ref{tab:3}, so that the two are not interchangeable.}
	\label{fig:13}
\end{figure}

The distribution of the node-level residuals is shown in Figure~\ref{fig:13}. The figure gives the node-wise absolute residual of the same case at four advancing frames and its variation along the depth, and gives a node-wise comparison of the predicted values with the reference values in the form of a scatter plot; the residual is more concentrated in the compression region near the projectile, and its magnitude decreases overall along the depth direction. The figure also gives the node-level coefficient of determination, whose node set and conditions of evaluation differ from those of the field-level relative $L_2$ error of Table~\ref{tab:3}, so that the two are not interchangeable.

\subsection{Derivation and reuse of field-level mechanism quantities}
\label{sec:6_6}

The operator outputs the complete tensor field of the six stress components on the mid-plane. The overwhelming majority of the quantities used in engineering to identify the failure state of concrete are the result of some criterion acting on the stress tensor. According to the information required, these criteria fall into two classes. The first class consists of algebraic functions of the stress state, such as the equivalent stress, the maximum principal stress, the hydrostatic pressure, the deviatoric stress invariants and Mohr-Coulomb type criteria; quantities of this class can be derived node by node directly from the six-component field and require no additional observable. The second class is integrated along the loading history, such as the equivalent plastic strain and the various damage variables; the stress history required for the integration is likewise available, because what the operator delivers is a sequence of fields in time. The six-component stress tensor field can therefore be used to derive the field-level criteria of the classes above.

The complete tensor information further makes a field-level discrimination of the failure mechanisms possible. The compression and shear comminution directly ahead of the projectile, the circumferential cracking in the tunnel region and the spalling at the rear face of the target are dominated by the compressive and the tensile stress states, respectively. Once the six-component field is decomposed into the hydrostatic pressure and the deviatoric part, the compression-dominated region corresponds to the high-value region of the pressure amplitude, whereas tensile failure corresponds to the region where the first principal stress is positive and the deviatoric stress is relatively high, so that the two mechanisms can be located separately on one and the same cross-section. This discrimination relies on all six components and cannot be given by a single scalar field.

What the operator delivers is a sequence of fields in time rather than a single-frame snapshot. Given three historical fields at equal intervals on one and the same case, the operator gives the prediction of the next frame at a fixed step. The independence of the criterion from the output target has a direct consequence: one and the same set of weights can simultaneously serve the computation of strength criteria, damage criteria and energy indicators, and when the criterion of engineering interest changes, the change occurs only in the post-processing stage. A model that takes some engineering quantity as its direct output target does not possess this property, and its criterion must be retrained as soon as it changes.

\subsection{Scope of applicability}
\label{sec:6_7}

The scope of applicability of the operator is defined by the conditions covered by its training and evaluation, comprising four items: the discretization, the input interface, the range of working conditions and the material parameters.

As regards the discretization, the solution adopts a hexahedral mesh with an edge length of 7.0 mm, and the field quantities are taken from the mid-plane of the target and resampled at a grid spacing of 7.8125 mm for storage as a regular grid. On this grid the operator works with the $32\times32$ node set obtained by sampling every other point, with a node spacing of 15.625 mm, and both the training and the evaluation are carried out on this node set.

As regards the input interface, the operator receives three mid-plane stress fields at equal intervals on one and the same case, together with the spatial coordinates of the nodes and the impact velocity; the historical frame interval is 16 frames and the prediction under the primary setting is single-step.

As regards the range of working conditions, the impact velocity takes the values 100, 135 and 165 m/s and the concrete compressive strength takes the values 14, 23 and 48 MPa, the two varying together across the three levels, so that the range of conditions covers 100 to 165 m/s and 14 to 48 MPa; the aggregates of the target are generated with a Fuller grading of 5 to 25 mm and a volume fraction of 42\%. The operators of the three velocity levels are trained independently, and the conclusions of the present work are limited to prediction within a single velocity level, with no claim of generalization of a single model across velocity levels.

As regards the material parameters, the material parameters of the projectile and of the target within the dataset take a fixed set of values, and the values of the material parameters themselves do not serve as input to the operator.

\section{Conclusions}
\label{sec:7}

For the task of time-varying prediction of the mid-plane six-component stress tensor field in concrete penetration, the physics-coupled graph neural operator (PC-GNO) is proposed in the present work, and field-level accuracy results are given under a cross-case setting. The task positioning differs from that of design filters that take macroscopic parameters as input: the input of the operator is the historical field produced during the solution process, which becomes available only after the computation, and its function is to give the prediction of subsequent fields without repeated solving.

The training data are generated by full-scale, aggregate-resolved mesoscale explicit finite element models, comprising 359 cases in total, divided by impact velocity into the three levels of 100, 135 and 165 m/s, which correspond respectively to the three concrete compressive strengths of 14, 23 and 48 MPa. The credibility of the data source is verified by laboratory ballistic penetration tests. The tests were carried out on the same ballistic apparatus against 23 MPa targets in two groups: in the first, penetration was performed with a projectile of full length 340 mm at an initial velocity of 134 m/s, and the measured deceleration duration is 2.51 ms; under the same projectile total length, impact velocity and target strength as this group the deceleration time history given by the simulation agrees with the measured oscillation curve in magnitude and in duration. In the second, penetration was performed with a projectile of full length 235 mm at an initial velocity of 135 m/s, and the crater depth measured by three-dimensional laser scanning is 133.0 $\pm$ 1.2 mm. The verification thus forms a two-step chain: the laboratory measurement data are used to verify the explicit finite element simulation and the data source, the simulation data are used to train the field-level operator, and the operator is then validated across cases.

Under the single-step primary setting, the average accuracy of the three levels is 80.16\%, the 100, 135 and 165 m/s levels giving 91.45\%, 77.80\% and 71.22\% respectively, the gains relative to the preceding-frame field of the same window being 27.81, 42.46 and 34.16 percentage points respectively. The number of learnable parameters of the operator is 20\,664. Under the same configuration, the same window set and the same starting protocol of the 100 m/s level, its accuracy is higher than the 87.9\% of the best configuration of the Fourier neural operator and the 71.0\% of the closed-form linear regression. The closed-form linear regression takes the same 22-dimensional node features and the same six output components as the operator; under the same window set and the same starting protocol, the accuracy of the operator is 20.45 percentage points higher than that of this linear reference.

What the operator outputs is the mid-plane six-component stress tensor field rather than some fixed engineering quantity, so that its use is not limited to a single criterion. Strength criteria, damage criteria and energy indicators can be derived from one and the same set of weights, and a change of criterion occurs only in the post-processing stage.

The contribution of the present work is threefold: the physics-coupled graph neural operator itself; an aggregate-resolved mid-plane six-component stress tensor dataset spanning three impact velocity levels and three concrete compressive strengths; and a two-step verification chain formed by the laboratory measurement data and the explicit finite element simulation.

\section*{CRediT authorship contribution statement}

\textbf{Wenpu Du}: Conceptualization, Methodology, Software, Validation, Formal analysis, Investigation, Data curation, Visualization, Writing -- original draft, Project administration.

\textbf{Peng Zhou}: Writing -- review \& editing, Resources.

\textbf{Yunlong Xia}: Writing -- review \& editing, Resources.

\textbf{Sinuo Xin}: Writing -- review \& editing, Resources.

\textbf{Congcong Zhang}: Writing -- review \& editing, Resources.

\textbf{Boyang Zhang}: Writing -- review \& editing, Resources.

\textbf{Yi Zhang}: Writing -- review \& editing, Resources.

\textbf{Wenzheng Xu}: Conceptualization, Writing -- review \& editing.

\section*{Declaration of competing interest}
The authors declare that they have no known competing financial interests or personal relationships that could have appeared to influence the work reported in this paper.

\section*{Funding}
This research did not receive any specific grant from funding agencies in the public, commercial, or not-for-profit sectors.

\end{document}